\pdfoutput=1

\RequirePackage{fix-cm}

\documentclass[a4paper]{article}

\usepackage[pdfpagemode={UseOutlines},bookmarks=true,bookmarksopen=true,
   bookmarksopenlevel=0,bookmarksnumbered=true,hypertexnames=false,
   colorlinks,linkcolor={blue},citecolor={blue},urlcolor={red},
   pdfstartview={FitV},unicode,breaklinks=true]{hyperref}

\hypersetup{
    colorlinks=false,
    pdfborder={0 0 0},
}

\usepackage{amssymb}
\usepackage{amsmath}

\usepackage{setspace} 		
\usepackage{lineno} 		

\usepackage{xcolor}

\usepackage{subfigure}

\newcommand{\citea}[1]{\cite{#1}}  
\newcommand{\citeb}[1]{\citep{#1}} 

\usepackage[acronym, nonumberlist, style=super]{glossaries}	
\glsdisablehyper
\usepackage{pgfplots}
\pgfplotsset{compat=newest}
\pgfplotsset{plot coordinates/math parser=false}

\newglossary{symbols}{sym}{sbl}{List of Symbols}
\newglossaryentry{greekletter}{name={\Large\textbf{\textsf{Greek Symbols}}},description={\nopostdesc}}
\newglossaryentry{romanletter}{name={\Large\textbf{\textsf{Roman Symbols}}},description={\nopostdesc}}

\newacronym{SRS}{SRS}{SIMONA Research Simulator}
\newacronym{DUT}{DUT}{Delft University of Technology}
\newacronym{CS}{C\&S}{Control \& Simulation}

\newacronym{DP}{DP}{Dynamic Programming}
\newacronym{MDP}{MDP}{Markov Decision Process}
\newacronym{ADP}{ADP}{Approximate Dynamic Programming}
\newacronym{RL}{RL}{Reinforcement Learning}
\newacronym{TD}{TD}{Temporal Difference}
\newacronym{NDP}{NDP}{Neuro Dynamic Programming}
\newacronym{SDP}{SDP}{Spline Dynamic Programming}
\newacronym{RLS}{RLS}{Recursive Least Squares}
\newacronym{LMS}{LMS}{Least Mean Square}

\newacronym{LS}{LS}{Least Squares}
\newacronym{WLS}{WLS}{Weighted Least-Squares}

\newacronym{gd}{GD}{Gradient Descent}
\newacronym{GD}{GD}{Gradient Descent}
\newacronym{nn}{NN}{Neural Networks}
\newacronym{mvsb}{MVSB}{Multivariate Simplex B-spline}
\newacronym{lti}{LTI}{Linear Time Invariant}

\newacronym{mil}{MIL}{Matrix Inversion Lemma}

\newacronym{RLSAPI}{RLSAPI}{Recursive Least Squares Approximate Policy Iteration}
\newacronym{RLSTD}{RLS TD}{Recursive Least Squares Temporal Difference}
\newacronym{MARS}{MARS}{Multivariate Adaptive Regression Splines}

\newglossaryentry{theta}{type=symbols,name=\ensuremath{\theta},sort=theta,description={Pole angle},parent=greekletter}
\newglossaryentry{thetad}{type=symbols,name=\ensuremath{\dot\theta},sort=thetad,description={Pole velocity},parent=greekletter}
\newglossaryentry{thetadd}{type=symbols,name=\ensuremath{\ddot\theta},sort=thetadd,description={Pole acceleration},parent=greekletter}
\newglossaryentry{s}{type=symbols,name=\ensuremath{s},sort=s,description={Distance from target},parent=romanletter}
\newglossaryentry{sd}{type=symbols,name=\ensuremath{\dot s},sort=sdot,description={Velocity cart},parent=romanletter}
\newglossaryentry{sdd}{type=symbols,name=\ensuremath{\ddot s},sort=sdotdot,description={Acceleration cart},parent=romanletter}
\newglossaryentry{g}{type=symbols,name=\ensuremath{g},description={Gravitational acceleration},sort=g,parent=romanletter}
\newglossaryentry{mc}{type=symbols,name=\ensuremath{m_c},sort=mc,description={Mass of cart},parent=romanletter}
\newglossaryentry{mp}{type=symbols,name=\ensuremath{m_p},sort=mp,description={Mass of pole},parent=romanletter}
\newglossaryentry{l}{type=symbols,name=\ensuremath{l},sort=l,description={Half length bar},parent=romanletter}
\newglossaryentry{t}{type=symbols,name=\ensuremath{t},sort=t,description={Time},parent=romanletter}
\newglossaryentry{x}{type=symbols,name=\ensuremath{\mathbf{x}},sort=x,description={State vector},parent=romanletter}
\newglossaryentry{xd}{type=symbols,name=\ensuremath{\mathbf{\dot x}},sort=xd,description={Time derivative of the state vector},parent=romanletter}

\newglossaryentry{u}{type=symbols,name=\ensuremath{\mathbf{u}},sort=u,description={Input},parent=romanletter}
\newglossaryentry{uj}{type=symbols,name=\ensuremath{u_j},sort=uj,description={Input scalar},parent=romanletter}
\newglossaryentry{cj}{type=symbols,name=\ensuremath{c_j},sort=cj,description={Input reward},parent=romanletter}
\newglossaryentry{U}{type=symbols,name=\ensuremath{U},sort=U,description={Valid input span},parent=romanletter}
\newglossaryentry{pi}{type=symbols,name=\ensuremath{\pi},sort=pi,description={Current policy},parent=greekletter}
\newglossaryentry{delta}{type=symbols,name=\ensuremath{\delta},sort=delta,description={Temporal difference error},parent=greekletter}
\newglossaryentry{z}{type=symbols,name=\ensuremath{\mathbf{z}},sort=z,description={Eligibility vector},parent=romanletter}
\newglossaryentry{r}{type=symbols,name=\ensuremath{r},sort=r,description={Reward},parent=romanletter}
\newglossaryentry{Rt}{type=symbols,name=\ensuremath{R_t},sort=Rk,description={Total reward},parent=romanletter}
\newglossaryentry{V}{type=symbols,name=\ensuremath{V},sort=V,description={True value}, parent=romanletter}
\newglossaryentry{Vt}{type=symbols,name=\ensuremath{\tilde  V},sort=Vt,description={Estimated value}, parent=romanletter}
\newglossaryentry{gamma}{type=symbols,name=\ensuremath{\gamma},sort=gamma,description={Discount factor},parent=greekletter}
\newglossaryentry{ac}{type=symbols,name=\ensuremath{\alpha_c},sort=alphac,description={Critic learning rate},parent=greekletter}
\newglossaryentry{aa}{type=symbols,name=\ensuremath{\alpha_a},sort=alphaa,description={Actor learning rate},parent=greekletter}
\newglossaryentry{lambda}{type=symbols,name=\ensuremath{\lambda},sort=lambda,description={Trace-decay parameter},parent=greekletter}
\newglossaryentry{q}{type=symbols,name=\ensuremath{q},sort=q,description={Number of steps from current position}, parent=romanletter}
\newglossaryentry{phi}{type=symbols,name=\ensuremath{\boldsymbol{\phi}},sort=phi,description={Generic parameter vector},parent=greekletter}

\newglossaryentry{I}{type=symbols,name=\ensuremath{\mathbf{I}},sort=I,description={Identity matrix},parent=romanletter}
\newglossaryentry{Nj}{type=symbols,name=\ensuremath{N_j},sort=Nj,description={Number of hidden neurons},parent=romanletter}
\newglossaryentry{W1b}{type=symbols,name=\ensuremath{\mathbf{\bar W_1}},sort=W1b,description={Input weights},parent=romanletter}
\newglossaryentry{W2b}{type=symbols,name=\ensuremath{\mathbf{\bar W_2}},sort=W2b,description={Output weights},parent=romanletter}
\newglossaryentry{W1}{type=symbols,name=\ensuremath{\mathbf{W_1}},sort=W1,description={Input weights, without bias},parent=romanletter}
\newglossaryentry{W2}{type=symbols,name=\ensuremath{\mathbf{W_2}},sort=W2,description={Output weights, without bias},parent=romanletter}

\newglossaryentry{fx}{type=symbols,name=\ensuremath{f(x)},sort=fx,description={Input layer function},parent=romanletter}
\newglossaryentry{hx}{type=symbols,name=\ensuremath{h(x)},sort=hx,description={Hidden layer function},parent=romanletter}
\newglossaryentry{gx}{type=symbols,name=\ensuremath{g(x)},sort=gx,description={Output layer function},parent=romanletter}

\newglossaryentry{e}{type=symbols,name=\ensuremath{\epsilon},sort=e,description={Difference in model- and system output},parent=romanletter}
\newglossaryentry{y}{type=symbols,name=\ensuremath{y},sort=y,description={Target value},parent=romanletter}
\newglossaryentry{yt}{type=symbols,name=\ensuremath{\tilde y},sort=yt,description={Model output},parent=romanletter}
\newglossaryentry{D1}{type=symbols,name=\ensuremath{\mathbf{D_1}},sort=D1,description={Derivative of hidden layer},parent=romanletter}

\newglossaryentry{b}{type=symbols,name=\ensuremath{\mathbf{b}},sort=b,description={Barycentric coordinate},parent=romanletter}
\newglossaryentry{upsilon}{type=symbols,name=\ensuremath{\upsilon},sort=upsilon,description={Vertex location},parent=greekletter}
\newglossaryentry{bern}{type=symbols,name=\ensuremath{B^d_\kappa},sort=bern,description={Bernstein basis polynomial},parent=romanletter}
\newglossaryentry{d}{type=symbols,name=\ensuremath{d},sort=d,description={Degree of spline},parent=romanletter}
\newglossaryentry{dh}{type=symbols,name=\ensuremath{\hat d},sort=d,description={Number of coefficients per simplex},parent=romanletter}
\newglossaryentry{ah}{type=symbols,name=\ensuremath{\hat a},sort=a,description={Total number of coefficients},parent=romanletter}
\newglossaryentry{kappa}{type=symbols,name=\ensuremath{\kappa},sort=kappa,description={Multi-index},parent=greekletter}
\newglossaryentry{p}{type=symbols,name=\ensuremath{p},sort=p,description={Evaluation of spline},parent=romanletter}
\newglossaryentry{ck}{type=symbols,name=\ensuremath{c_\kappa},sort=ckappa,description={B-coefficient},parent=romanletter}
\newglossaryentry{Ch}{type=symbols,name=\ensuremath{\mathbf{c}},sort=C,description={Estimated global vector of B-coefficients},parent=romanletter}
\newglossaryentry{BERN}{type=symbols,name=\ensuremath{\mathbf{B}^d},sort=BERN,description={Lexicographically sorted vector of Bernstein basis polynomials},parent=romanletter}
\newglossaryentry{B}{type=symbols,name=\ensuremath{\mathbf{B}},sort=B,description={Lexicographically sorted vector of Bernstein basis polynomials},parent=romanletter}
\newglossaryentry{D}{type=symbols,name=\ensuremath{\mathbf{D}},sort=D,description={The diagonal data membership matrix},parent=romanletter}
\newglossaryentry{X}{type=symbols,name=\ensuremath{X},sort=X,description={Linear regression matrix},parent=romanletter}
\newglossaryentry{Y}{type=symbols,name=\ensuremath{\mathbf{Y}},sort=Y,description={Target output vector},parent=romanletter}
\newglossaryentry{R}{type=symbols,name=\ensuremath{\mathbf{r}},sort=r,description={Residual vector},parent=romanletter}
\newglossaryentry{COST}{type=symbols,name=\ensuremath{\mathcal{J}},sort=J,description={Cost function},parent=romanletter}
\newglossaryentry{Sigma}{type=symbols,name=\ensuremath{\mathbf{\Sigma}},sort=Sigma,description={Residual covariance matrix},parent=greekletter}
\newglossaryentry{nu}{type=symbols,name=\ensuremath{\boldsymbol{\nu}},sort=nu,description={Vector of Lagrange multipliers},parent=greekletter}
\newglossaryentry{nuh}{type=symbols,name=\ensuremath{\boldsymbol{\hat \nu}},sort=nuh,description={Estimate vector of Lagrange multipliers},parent=greekletter}
\newglossaryentry{H}{type=symbols,name=\ensuremath{\mathbf{H}},sort=H,description={The smoothness matrix},parent=romanletter}
\newglossaryentry{xb}{type=symbols,name=\gls{B},sort=xb,description={Regression vector},parent=romanletter}
\newglossaryentry{P}{type=symbols,name=\ensuremath{\mathbf{P}},sort=P,description={Inverse of estimated parameter covariance matrix},parent=romanletter}
\newglossaryentry{W}{type=symbols,name=\ensuremath{\mathbf{W}},sort=W,description={Influence matrix},parent=romanletter}
\newglossaryentry{Z}{type=symbols,name=\ensuremath{\mathbf{Z}},sort=Z,description={Orthogonal projector},parent=romanletter}
\newglossaryentry{L}{type=symbols,name=\ensuremath{\mathbf{L}},sort=L,description={Kalman gain},parent=romanletter}
\newglossaryentry{Im}{type=symbols,name=\ensuremath{\mathbf{I}_m},sort=Im,description={Identity matrix of size m},parent=romanletter}
\newglossaryentry{sigmar}{type=symbols,name=\ensuremath{\sigma_r},sort=sigmar,description={Covariance of residue},parent=greekletter}
\newglossaryentry{forget}{type=symbols,name=\ensuremath{\beta},sort=beta,description={Forget factor},parent=greekletter}
\newglossaryentry{forgetV}{type=symbols,name=\ensuremath{\boldsymbol{\lambda}},sort=lambdaV,description={Forget factor vector},parent=greekletter}
\newglossaryentry{T}{type=symbols,name=\ensuremath{T},sort=T,description={Total number of simplices in the triangulation},parent=romanletter}
\newglossaryentry{Chp}{type=symbols,name=\ensuremath{\mathbf{\hat c}_P},sort=Chp,description={Estimated global vector of B-coefficients projected on the basis P},parent=romanletter}
\newglossaryentry{gain}{type=symbols,name=\ensuremath{K},sort=K,description={Adjustment gain},parent=romanletter}
\newglossaryentry{K1}{type=symbols,name=\ensuremath{\mathbf{K}_1},sort=K1,description={Smoothness Direction Matrix},parent=romanletter}
\newglossaryentry{hh}{type=symbols,name=\ensuremath{\mathbf{\hat h}},sort=hh,description={Smoothness offset vector},parent=romanletter}
\newglossaryentry{eta}{type=symbols,name=\ensuremath{\eta},sort=eta,description={Step-size},parent=greekletter}
\newglossaryentry{tau}{type=symbols,name=\ensuremath{\tau},sort=tau,description={Forget factor equation parameter},parent=greekletter}

\newglossaryentry{sigma2}{type=symbols,name=\ensuremath{\sigma^2},sort=sigma,description={Standard deviation},parent=greekletter}
\newglossaryentry{mu}{type=symbols,name=\ensuremath{\mu},sort=mu,description={Mean value},parent=greekletter}

\makeglossaries

\graphicspath{ {./pics/} }

\usepackage{natbib}

\usepackage[figwidth=5cm]{todonotes}

\usepackage[]{cleveref}

\newenvironment{definition}[1][Definition]{\begin{trivlist}
\item[\hskip \labelsep {\bfseries #1}]}{\end{trivlist}}

\title{On Approximate Dynamic Programming with Multivariate Splines for Adaptive Control}
\author{W.J. Eerland\footnote{University of Southampton, Transportation Research Group  [\texttt{w.j.eerland@soton.ac.uk}]}, C.C. de Visser\footnote{Delft University of Technology, Control and Simulation}, E. van Kampen\footnote{Delft University of Technology, Control and Simulation}}
\date{}

\begin{document}


\maketitle


\begin{abstract}
	We define a \acrshort{SDP} framework based on the \acrshort{RLSTD} algorithm and multivariate simplex B-splines. We introduce a local forget factor capable of preserving the continuity of the simplex splines. This local forget factor is integrated with the \acrshort{RLSTD} algorithm, resulting in a modified \acrshort{RLSTD} algorithm that is capable of tracking time-varying systems. We present the results of two numerical experiments, one validating \acrshort{SDP} and comparing it with \acrshort{NDP} and another to show the advantages of the modified \acrshort{RLSTD} algorithm over the original. While \acrshort{SDP} requires more computations per time-step, the experiment shows that for the same amount of function approximator parameters, there is an increase in performance in terms of stability and learning rate compared to \acrshort{NDP}. The second experiment shows that \acrshort{SDP} in combination with the modified \acrshort{RLSTD} algorithm allows for faster recovery compared to the original \acrshort{RLSTD} algorithm when system parameters are altered, paving the way for an adaptive high-performance non-linear control method.
\end{abstract}


\glsresetall 



\section{Introduction}
Bellman described multi-stage decision processes from a mathematical point of view in \citea{Bellman1957}, this algorithm was called \gls{DP}. The \gls{DP} principle comes down to describing each state with a value in a value function and moving the system to the state with the highest value. 
The value function is also called the cost-to-go function and stores the expected sum of future rewards for each state. In general this function cannot be found directly, therefore an iterative approach like \gls{TD}-learning \citeb{SB98, Sut88} is used.

\gls{DP} has the ability to solve complex control problems in diverse environments, and it is possible to view the environment as a black-box by modelling it using system identification techniques. The advantage of combining \gls{DP} and system identification techniques, is that it leads to an adaptive control scheme.
These adaptive controllers have already been successfully trained off-line for many purposes, ranging from agile missile interception \citeb{HB99} to aircraft auto-landing and control \citeb{SB97}. Also, an on-line adaptive critic flight control was implemented on a six-degree-of-freedom business jet aircraft over its full operating envelope, improving its performance when unexpected conditions are encountered for the first time \citeb{FS04}.

Barto, Sutton and Anderson used neural networks to parametrise the value function \citeb{BSA83}. However, this function approximator is non-linear in the parameters, with the result that stability can only be guaranteed if bounded network weights are used, where bounds are determined by off-line analysis. This approach has been applied to examples \citeb{HB99,SB97,FS04} mentioned before.
More recently, an adaptive controller has been introduced in \citea{AYB+07}, and again the neural network weights are bounded to guarantee stability. A drawback is that for a time-varying system these bounds shift and stability can no longer be guaranteed. This combination of neural networks and \gls{DP} is commonly referred to as \gls{NDP} \citeb{bertsekas1996neuro}.

There is a proof of convergence when linear-in-the-parameters function approximators are used in \citea{TVR97}, however, this proof demands knowledge of the shape of the optimal value function. 
With the \gls{RLSTD} algorithm convergence in a stochastic framework is assured \citeb{BB96}, even when the linear regression basis cannot perfectly fit the value function.
In the last decade, the \gls{DP} theory in continuous time and space has been further developed in \citea{Doy96, Doy00, MD05}. 
More recently, the proof of convergence has been extended to include the optimal policy \citeb{ma2009}, however one of the problems that remains is the a-priori unknown shape of the value function.

Using polynomials as a function approximator in a \gls{DP} framework was investigated by Bellman in 1963 \citeb{BKK63}. In 1985, \citea{Sch85} concentrates on the use of global polynomials in combination with \gls{DP}. 
With the development of the \gls{RLSTD} algorithm, it is possible to obtain a proven convergence by combining it with a polynomial approximation as discussed in \citea{ma2009}. However the limitation is that the approximation power of global polynomials can only be increased by increasing the order of the polynomials, which will also lead to numerical instabilities in the solution schemes of the approximation. According to \citea{summers2013}, using the sum of squares allows the use of higher order polynomials, but will eventually still lead to numerical instability.

A recommendation in \citea{ma2009} is to use local polynomial regression to treat \gls{MDP} problems with value functions of unknown form. This recommendation is supported by \citea{Da76}, which states that it is highly desirable from an efficiency point of view to use local polynomial regression. Recently a novel method based on multivariate simplex B-splines has been applied in a linear regression framework \citeb{deVisser2009a}. The use of a local polynomial basis allows for transparency and efficient (sparse) computational schemes. Furthermore, the spatial location of the B-coefficients and modularity of the triangulation allow for local model modification and refinement \citeb{Lai2007b}. These properties make multivariate simplex B-splines an excellent candidate for use in the \gls{DP} framework.

There are existing approaches which combines \gls{MARS} \citeb{Friedman1991} and \gls{DP} \citeb{Chen1999,Chen19992,Cervellera2007}. However, multivariate simplex B-splines distinguishes itself from \gls{MARS} in terms of computational efficiency by using B-splines \citeb{Bakin2000}. And they are supported by a triangulation of simplices, allowing functionality in a non-square domain \citeb{Lai2007b}.

The contribution of this paper is a framework that allows the use of multivariate simplex B-splines in combination with the \gls{RLSTD} algorithm, giving rise to \gls{SDP} that enables control of non-linear stochastic systems. A method for continuous local value function adaptation is presented which is enabled by the spatial location property of the coefficients of the multivariate splines; this is achieved by implementing a new formulation for the covariance update step. The effectiveness of this \gls{SDP} framework is investigated by comparing it with \gls{NDP} in terms of computational complexity and performance. Furthermore, the \gls{RLSTD} algorithm is modified to allow for adaptive control of time-varying systems. The validation and comparison of both cases are investigated with a non-linear 2D control problem, the pendulum swing-up.

In section \ref{sec:dp}, we briefly introduce the \gls{DP} framework and show how the value function and greedy policy are constructed. In section \ref{sec:splines}, we give a brief introduction on the mathematical background of multivariate simplex B-splines, the function approximator used in the \gls{SDP} framework. The \gls{SDP} framework itself is explained in section \ref{sec:sdp}; the main purpose of this section is to address the steps specific to using multivariate simplex B-splines for value function approximation in combination with the \gls{RLSTD} algorithm. Section \ref{sec:results} is there to demonstrate that the \gls{SDP} framework indeed works for the given control problem. \gls{SDP} is compared with neural networks for system performance on a stochastic system and a time-varying system. These results are discussed in section \ref{sec:discussion} and finally conclusions and recommendations are presented in section \ref{sec:conclusion}.

\section{Preliminaries on Dynamic Programming} \label{sec:dp}
In this section, we present the preliminaries on \gls{DP}, the algorithm that is part of the \gls{SDP} framework.
For a more complete description we refer to \citea{SB98, SBP+04, WBP07, BBDS+10, Ber07}.
We start with a brief overview of the \gls{MDP} followed by the policy evaluation and concluded with the policy improvement.

\subsection{Markov Decision Processes}
The policy evaluation problem associated with discrete-time stochastic optimal control problems is referred to as a \gls{MDP}. Finding the solution to an \gls{MDP} is a sequential optimization problem where the goal is to find a policy that maximizes the sum of the expected infinite-horizon discounted rewards.

Let $\gls{x}_t \in X$ be the state vector and $\gls{u}_t \in U$ be the input vector, both at time $t$, where $\gls{u}_t$ is determined by the policy $\gls{pi}$ and $X$ and $U$ denote the finite sets of states and inputs. The reward function is $\gls{r}_{t+1}(\gls{x}_{t+1},\gls{u}_t)$ and $0\leq\gls{gamma}<1$ is the discount factor. The goal is to find a policy $\pi$ that obtains the maximum total reward.

For each policy $\gls{pi}$ there exists a value function $V^{\gls{pi}}(\gls{x}_t)$ that indicates a measure of long-term performance at each state:
\begin{equation} V^{\gls{pi}}(\gls{x}_t) = \sum_{k=t}^{\infty} \gls{gamma}^{k-t} \gls{r}_{k+1}(\gls{x}_{k+1},\gls{u}_k) \label{eq:valuefunc} \end{equation}
The objective can now be formulated as finding a policy $\gls{pi}^*$ such that $V^{\gls{pi}^*}(\gls{x}) \geq V^{\gls{pi}}(\gls{x})$ for all $\gls{x} \in X$ and for all policies $\gls{pi}$. This policy $\gls{pi}^*$ is called the optimal policy and can be found by applying both policy evaluation and policy improvement \citea{SB98}.

The policy evaluation determines the $V^{\gls{pi}}(\gls{x}_t)$ of the current policy $\gls{pi}$, where the policy improvement uses this knowledge to adjust the policy $\gls{pi}$ such that it ends up in the most valuable states.

\subsection{Policy evaluation} \label{sec:policyeval}
In order to evaluate the value function in an iterative fashion, \citea{Sut88} uses \gls{TD}-learning. For \gls{TD}-learning, the following must hold:
\begin{equation} \begin{array}{rl}
V^{\gls{pi}}(\gls{x}_t) = 	& \gls{r}_{t+1} + \sum_{k={t+1}}^{\infty} \gls{gamma}^{k-t} \gls{r}_{k+1} \\
= 				& \gls{r}_{t+1} + \gls{gamma} V^{\gls{pi}}(\gls{x}_{t+1})
\end{array} \end{equation}
If this equality does not hold, the difference is called the \gls{TD}-error:
\begin{equation} e_t = \gls{r}_{t+1} + \gls{gamma} V^{\gls{pi}}(\gls{x}_{t+1}) - V^{\gls{pi}}(\gls{x}_t) \label{eq:td} \end{equation}
By minimizing this \gls{TD}-error, the value function can be constructed in an iterative approach.
In order to construct a value function for a continuous problem, there is need for parameterization to describe a complete state space with a  finite number of parameters.
The value function now becomes $V^{\gls{pi}}(\gls{x}_t,\gls{Ch}_t)$, where $\gls{Ch}_t$ are the parameters at time $t$ that shape the continuous value function in domain $X$.

In \citea{BB96} the \gls{RLSTD} algorithm was introduced; this algorithm and its computational complexity is visible in Table~\ref{tab:trainRLSTD}. Here $\gls{dh}$ and $\gls{ah}$ represent the number of coefficients per simplex and total number of coefficients respectively, $\gls{xb}$ represents the linear regression matrix and \gls{P} is the parameter covariance matrix. These parameters will be further explained in section~\ref{sec:splines}. \gls{RLSTD} converges to the least squares approximation of the optimal value function $\gls{Ch}^*$, given that each state $\gls{x} \in \gls{X}$ is visited infinitely often.
Although \gls{RLSTD} requires more computations per time-step than \acrshort{TD}($\lambda$) algorithms \citeb{Sut88}, it is more efficient in the statistical sense as more information is extracted from training experience, allowing it to converge faster \citeb{BB96}. Furthermore, while \gls{LMS} aims to decrease the mean square error $e_t$ at each time-step separately, \gls{RLSTD} minimizes this objective function:
\begin{equation} \gls{COST}_t = \frac{1}{t} \sum_{k=0}^{t} \left[ e_k \right]^2 \label{eq:classicLS} \end{equation}

\begin{table}
\caption{\acrshort{RLSTD} algorithm, from \protect{\citea{BB96}}}
\label{tab:trainRLSTD}
\centering
 \begin{tabular}{lrll}
  \hline
  \hline
	  \textbf{Step} & & \textbf{Action} & \textbf{Computational} \\
			& & & \textbf{Complexity} 		\\
    \hline
(1)& $e_t =$ & $\gls{r}_{t+1} - (\gls{xb}_t - \gls{gamma} \gls{xb}_{t+1})^{\top} \gls{Ch}_{t}$ & $\mathcal{O}(\gls{dh})$ \\
(2)& $\gls{P}_{t+1} =$ & $\gls{P}_{t} -  \frac{ \gls{P}_{t} \gls{xb}_t ( \gls{xb}_t - \gls{gamma} \gls{xb}_{t+1} )^{\top} \gls{P}_{t}}{1 + ( \gls{xb}_t - \gls{gamma} \gls{xb}_{t+1} )^{\top} \gls{P}_{t} \gls{xb}_t }  $ & $\mathcal{O}(\gls{ah}^2)$ \\
(3) & $ \gls{Ch}_{t+1} =$ & $\gls{Ch}_{t} +  \frac{ \gls{P}_{t} }{ 1 + ( \gls{xb}_t - \gls{gamma} \gls{xb}_{t+1} )^{\top} \gls{P}_{t} \gls{xb}_t }  \gls{xb}_t e_t$ & $\mathcal{O}(\gls{ah}^2)$ \\
  \hline
  \hline
 \end{tabular}
\end{table}

\subsection{Policy improvement} \label{sec:pi}
The computation of the value function is called policy evaluation. Using this value function, a greedy action can be selected. If a policy is updated in this manner,  it is called (greedy) policy improvement:
\begin{equation} \gls{u}_t ( \gls{x}_t ) = \max_{\gls{u}_t \in \gls{U}} \left[ \gls{V}^{\gls{pi} } ( \gls{x}_t )  \right] \end{equation}
The repetition of the policy evaluation and policy improvement is called policy iteration and will result in an optimal policy \citeb{SB98}. According to \citea{Doy96}, the optimal non-linear feedback control law is a function of value function's gradient:
\begin{equation} \gls{u}_t ( \gls{x}_t ) = \gls{u}^{\max} ~ g \left( \frac{1}{ c } \gls{tau} \frac{\partial \gls{V}^{\gls{pi}} ( \gls{x}_t ) }{\partial \gls{x}_t} \frac{\partial f(\gls{x}_t,\gls{u}_t)}{\partial \gls{u}_t} \right) \end{equation}
where $\gls{u}^{\max}$ is the maximum control input, $g(x) = \tanh (\frac{\pi}{2} x)$, $c$ is the control cost parameter, $\tau$ is the step-size parameter and $f(\gls{x}_t,\gls{u}_t)$ are the system dynamics.

This optimal control law is applied to the pendulum swing-up task, with the result visible in \ref{sec:swingup}. Having discussed both the policy evaluation and improvement, section~\ref{sec:splines} will now describe the parametrization of the value function using multivariate simplex B-splines.

\section{Preliminaries on Multivariate Simplex B-Splines} \label{sec:splines}
This section serves as a brief introduction to the mathematical theory of the simplex B-splines. 
For a more extensive and general introduction to multivariate spline theory we refer to \citea{Lai2007b}.
We start by introducing the basic concept of a single basis polynomial and B-form, then introduce the triangulation, followed by the vector notation of the B-form. Finally the \gls{RLS} estimator for simplex splines is reviewed.

\subsection{Simplex and barycentric coordinates}
The polynomial basis of a multivariate simplex B-spline is defined on a simplex. A simplex is defined by the non-degenerate vertices $ ( \gls{upsilon}_0, \gls{upsilon}_1, \ldots , \gls{upsilon}_n ) \in \mathbb{R}^n $ and thus creates a span in $n$-dimensional space. Any point $ \gls{x} = ( x_1 , x_2 , \ldots , x_n ) $ can be transformed to a barycentric coordinate $ \gls{b}(\gls{x}) = ( {b}_0 , {b}_1 , \ldots , {b}_n ) $ with respect to a simplex. 
The relation between Cartesian coordinate $\gls{x}$ and barycentric coordinate $\gls{b}(\gls{x})$ is:
\begin{equation} \gls{x} = \sum^n_{i=0} b_i \gls{upsilon}_i \qquad \sum^n_{i=0} b_i = 1 \end{equation}

\subsection{Triangulation}
Any number of simplices can be combined into a triangulation, where
a triangulation $\mathcal{T}$ is a special partitioning of a domain into a set of $\mathit{J}$ non-overlapping simplices and is defined in \citea{Lai2007b} as:
\begin{equation} \mathcal{T} \equiv \bigcup_{i=1}^{\mathit{J}} t_i , \quad t_i \cap t_j \in \{ \emptyset, \tilde t \}, \quad \forall t_i, t_j \in \mathcal{T} \quad , i \neq j \end{equation}
with $\tilde t$ a simplex of dimension $(<n)$.
A popular triangulation method is Delaunay triangulation \citeb{Lee1980}. Recently, a new method for creating globally optimal triangulations named Intersplines was introduced in \citea{dVvK+12}.

\subsection{Basis functions}
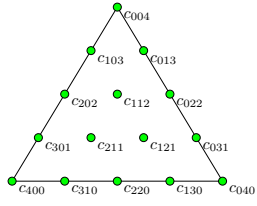
\begin{figure}
\centering
\resizebox{!}{3cm}{
%
%
%
\begin{tikzpicture}

\begin{axis}[%
width=6cm,
height=5cm,
scale only axis,
xmin=-0.1,
xmax=2.5,
ymin=-0.2,
ymax=1.1,
hide axis,
axis x line*=bottom,
axis y line*=left
]
\addplot [
color=black,
mark size=2.5pt,
only marks,
mark=*,
mark options={solid,fill=green,draw=black},
forget plot
]
table[row sep=crcr]{
0 0\\
};
\node[right, inner sep=0mm, text=black]
at (axis cs:0.06, -0.05) {$c_{400}$};
\addplot [
color=black,
mark size=2.5pt,
only marks,
mark=*,
mark options={solid,fill=green,draw=black},
forget plot
]
table[row sep=crcr]{
0.5 0\\
};
\node[right, inner sep=0mm, text=black]
at (axis cs:0.56, -0.05) {$c_{310}$};
\addplot [
color=black,
mark size=2.5pt,
only marks,
mark=*,
mark options={solid,fill=green,draw=black},
forget plot
]
table[row sep=crcr]{
0.25 0.25\\
};
\node[right, inner sep=0mm, text=black]
at (axis cs:0.31, 0.2) {$c_{301}$};
\addplot [
color=black,
mark size=2.5pt,
only marks,
mark=*,
mark options={solid,fill=green,draw=black},
forget plot
]
table[row sep=crcr]{
1 0\\
};
\node[right, inner sep=0mm, text=black]
at (axis cs:1.06, -0.05) {$c_{220}$};
\addplot [
color=black,
mark size=2.5pt,
only marks,
mark=*,
mark options={solid,fill=green,draw=black},
forget plot
]
table[row sep=crcr]{
0.75 0.25\\
};
\node[right, inner sep=0mm, text=black]
at (axis cs:0.81, 0.2) {$c_{211}$};
\addplot [
color=black,
mark size=2.5pt,
only marks,
mark=*,
mark options={solid,fill=green,draw=black},
forget plot
]
table[row sep=crcr]{
0.5 0.5\\
};
\node[right, inner sep=0mm, text=black]
at (axis cs:0.56, 0.45) {$c_{202}$};
\addplot [
color=black,
mark size=2.5pt,
only marks,
mark=*,
mark options={solid,fill=green,draw=black},
forget plot
]
table[row sep=crcr]{
1.5 0\\
};
\node[right, inner sep=0mm, text=black]
at (axis cs:1.56, -0.05) {$c_{130}$};
\addplot [
color=black,
mark size=2.5pt,
only marks,
mark=*,
mark options={solid,fill=green,draw=black},
forget plot
]
table[row sep=crcr]{
1.25 0.25\\
};
\node[right, inner sep=0mm, text=black]
at (axis cs:1.31, 0.2) {$c_{121}$};
\addplot [
color=black,
mark size=2.5pt,
only marks,
mark=*,
mark options={solid,fill=green,draw=black},
forget plot
]
table[row sep=crcr]{
1 0.5\\
};
\node[right, inner sep=0mm, text=black]
at (axis cs:1.06, 0.45) {$c_{112}$};
\addplot [
color=black,
mark size=2.5pt,
only marks,
mark=*,
mark options={solid,fill=green,draw=black},
forget plot
]
table[row sep=crcr]{
0.75 0.75\\
};
\node[right, inner sep=0mm, text=black]
at (axis cs:0.81, 0.7) {$c_{103}$};
\addplot [
color=black,
mark size=2.5pt,
only marks,
mark=*,
mark options={solid,fill=green,draw=black},
forget plot
]
table[row sep=crcr]{
2 0\\
};
\node[right, inner sep=0mm, text=black]
at (axis cs:2.06, -0.05) {$c_{040}$};
\addplot [
color=black,
mark size=2.5pt,
only marks,
mark=*,
mark options={solid,fill=green,draw=black},
forget plot
]
table[row sep=crcr]{
1.75 0.25\\
};
\node[right, inner sep=0mm, text=black]
at (axis cs:1.81, 0.2) {$c_{031}$};
\addplot [
color=black,
mark size=2.5pt,
only marks,
mark=*,
mark options={solid,fill=green,draw=black},
forget plot
]
table[row sep=crcr]{
1.5 0.5\\
};
\node[right, inner sep=0mm, text=black]
at (axis cs:1.56, 0.45) {$c_{022}$};
\addplot [
color=black,
mark size=2.5pt,
only marks,
mark=*,
mark options={solid,fill=green,draw=black},
forget plot
]
table[row sep=crcr]{
1.25 0.75\\
};
\node[right, inner sep=0mm, text=black]
at (axis cs:1.31, 0.7) {$c_{013}$};
\addplot [
color=black,
mark size=2.5pt,
only marks,
mark=*,
mark options={solid,fill=green,draw=black},
forget plot
]
table[row sep=crcr]{
1 1\\
};
\node[right, inner sep=0mm, text=black]
at (axis cs:1.06, 0.95) {$c_{004}$};
\addplot [
color=black,
solid,
forget plot
]
table[row sep=crcr]{
0 0\\
2 0\\
1 1\\
0 0\\
};
\end{axis}
\end{tikzpicture}%
}
\caption{B-net overview of a $4^{th}$ degree simplex spline}
\label{fig:bnet}
\end{figure}

The polynomial basis of the simplex spline are the Bernstein basis polynomials $\gls{bern} (\gls{b})$, where $d$ is the degree of the spline and $\gls{b}$ is the barycentric coordinate discussed earlier:
\begin{equation} \gls{bern} (\gls{b}) = \frac{d!}{\kappa !} \gls{b}^\kappa \label{eq:bern}\end{equation}
here $\kappa \geq 0$ is a multi-index, which has properties: $\kappa ! = \kappa_0 ! \kappa_1 ! \cdots \kappa_n !$, $| \kappa | = \kappa_0 + \kappa_1 + \cdots + \kappa_n$ and $\gls{b}^\kappa = b_0^{\kappa_0} b_1^{\kappa_1} \cdots b_n^{\kappa_n} $. The valid permutations of $\kappa$ with the constraint $| \kappa | = d$ equal the total number of B-coefficients and basis polynomials per simplex and is equal to:
\begin{equation} \gls{dh} = \frac{(d+n)!}{n!d!} \end{equation}
In combination with a set of $\mathit{J}$ non-overlapping simplices, the total number of B-coefficients for a complete triangulation is:
\begin{equation} \gls{ah} = \mathit{J} \cdot \gls{dh} \end{equation}
The multi-index $\kappa$ has a requirement on the ordering, called a lexicographical sorting order which is introduced in \citea{Lai2007b}. This means $\kappa_{\nu \mu \varkappa}$ comes before $\kappa_{i j k}$ provided that $\nu > i$, or if $\nu = i$, then $\mu > j$, or if $\nu = i$ and $\mu = j$, then $\varkappa > k$. Thus for $d=2$ the order is:
\begin{equation}  \kappa_{2,0,0} , \quad \kappa_{1,1,0} , \quad \kappa_{1,0,1} , \quad \kappa_{0,2,0} , \quad \kappa_{0,1,1} , \quad \kappa_{0,0,2} \label{eq:lex} \end{equation}
Each B-coefficient has a unique position within the simplex based on $\gls{kappa}$; this relation between B-coefficient and spatial position allows the creation of a B-net as seen in Figure~\ref{fig:bnet}.


\subsection{Vector formulation of the B-form}
In order to complete the vector formulation, $\gls{BERN}_{t_j} (\gls{b}) \in \mathbb{R}^{\hat{d} \times 1}$ is introduced as a vector of Bernstein basis polynomials (Eq. \ref{eq:bern}) of simplex $t_j$, which are sorted lexicographically as indicated by Eq.~\ref{eq:lex}. Adapted from \citea{visser2013}, we define the B-form as a row vector on simplex $t_j$:
\begin{equation}  \gls{p}(\gls{b}) = \left\{ \begin{array}{rl} \gls{BERN}(\gls{b})^{\top} \cdot \gls{Ch}^{t_j} ,& \gls{x} \in t_j \\ 0 ,& \gls{x} \notin t_j \end{array} \right. \end{equation}
where $\mathbf{c}^{t_j}$ are the B-coefficients on simplex $t_j$. 
The matrix operation to evaluate the simplex B-spline function of degree $d$ and continuity order $r$, defined on a triangulation $\mathcal{T}_\mathit{J}$ is:
\begin{equation}
 s_d^r (\gls{b}) \equiv \gls{B}(\gls{b})^{\top} \cdot \gls{Ch} \in \mathbb{R}, \quad \gls{x} \in \mathcal{T}_\mathit{J}
\end{equation}
Now $\gls{B}(\gls{b})$ (note the absence of the superscript 'd') is the global vector of basis polynomials:
\begin{equation} \gls{B}(\gls{b}) \equiv \left[ \gls{BERN}(\gls{b})_{t_1}^{\top} ~ \gls{BERN}(\gls{b})_{t_2}^{\top} ~ \cdots ~ \gls{BERN}(\gls{b})_{t_J}^{\top} \right]^{\top} \in \mathbb{R}^{\gls{ah} ~ \mathtt{x} ~  1} \end{equation}
The global vector of B-coefficients \gls{Ch} is defined as:
\begin{equation} \gls{Ch} \equiv \left[ {\gls{Ch}^{t_1}}^{\top} ~ {\gls{Ch}^{t_2}}^{\top} ~ \cdots ~ {\gls{Ch}^{t_J}}^{\top} \right]^{\top} \in \mathbb{R}^{\gls{ah} ~ \mathtt{x} ~  1} \label{eq:defc} \end{equation}
The spline space is the space of all spline functions $s_d^r$ in the triangulation $\mathcal{T}$. We use the definition of the spline space from \citea{Lai2007b}:
\begin{equation} S_d^r (\mathcal{T}) \equiv \{ s_d^r \in C^r(\mathcal{T}) : s_d^r |_t \in \mathbb{P}_d , \forall t \in \mathcal{T} \} \end{equation}
where $\mathbb{P}_d$ is the space of polynomials of degree $d$.

\subsection{Continuity} \label{sec:continuity}
Since $\gls{p}(\gls{b})$ is a linear combination of continuous functions, $s_d^r (\gls{b})$ is naturally continuous on each simplex. However, in order to assure continuity of $S_d^r (\mathcal{T})$ between simplices, constraints are imposed on the relations between the coefficients of different simplices. The continuity order $r$ fixes the derivatives $\frac{d^r p}{db^r}$ on the edges between neighboring simplices. The required continuity conditions can be calculated using \citea{Lai2007b}:
\begin{equation} \begin{array}{r}
c^{t_i}_{(\kappa_0,\ldots,\kappa_{n-1},m)} = \sum_{| \gamma | = m} c^{t_j}_{(\kappa_0,\ldots,\kappa_{n-1},0) + \gamma} B^m_\gamma (w) 
\\ \\
0 \leq m \leq r
\end{array} \label{eq:contcond} \end{equation}
Here $w$ is a vertex of simplex $t_j$ which is not found on the edge that is shared with simplex $t_i$. 
All constraints required for continuity are collected in the smoothness matrix $\gls{H}$, with each row containing a new constraint and the columns consisting of the coefficients \citea{deVisser2009a,deVisser2011}. These equations are all equaled to zero, resulting in the following matrix form:
\begin{equation}  \gls{H} \gls{Ch} = 0 \label{eq:matrixH} \end{equation}
with \gls{Ch} as in Eq.~\ref{eq:defc}.

\subsection{Approximation power}
To describe the approximation power, the following definition from \citea{Lai2007b}, chapter $10.1$ is used:
\begin{definition} \label{def:approx}
\emph{(Approximation power of $S_d^r (\mathcal{T})$)} 
Fix $0 \leq r < d$ and $0 < \theta \leq \pi /3$. Let $m$ be the largest integer such that for every polygonal domain $\Omega$ and every regular triangulation $\mathcal{T}$ of $\Omega$ with smallest angle $\theta$, for every $f \in W_q^m ( \Omega )$, there exists a spline $s \in S_d^r (\mathcal{T})$ with
\begin{equation} ||f - s||_{q,\Omega} \leq K ~ |\mathcal{T}|^m ~ |f|_{m,q,\Omega} \label{eq:approx} \end{equation}
where the constant $K$ depends only on $r$, $d$, $\theta$, and the Lipschitz constant of the boundary of $\Omega$. Then we say that $S_d^r$ has approximation power $m$ in the $q$-norm. If this holds for $m = d + 1$, we say that $S_d^r$ has full approximation power in the $q$-norm.
\end{definition}
The theory behind this is extensive and available in \citea{Lai2007b}, however for now it is important to realize that $|\mathcal{T}|^m$ is a function of the longest edge in triangulation $\mathcal{T}$. This reveals that it is possible to locally increase the approximation power by reducing the length of the longest edge in $\mathcal{T}$. 

\subsection{Recursive least squares} \label{sec:rls}
\gls{RLS} is a method which allows the estimated parameters $\gls{Ch}$ to be updated online with the use of the parameter covariance matrix $\gls{P}$ \citea{deVisser2011}.
The algorithm and the computational complexity is found in Table~\ref{tab:trainMIL}. 
Note that it is essential to keep the column relations of \gls{P} intact to enforce the constraints $\gls{H} \gls{Ch} = 0$ from Eq.~\ref{eq:matrixH}.
To initialize the \gls{P} matrix, we use $\mathbf{Z}$, the orthogonal projector on the null-space of \gls{H} \citea{Lawson1974}:
\begin{equation} \gls{P}_1 = \gls{forget}_1 ~ \mathbf{Z} \label{eq:p0} \end{equation}
where $\mathbf{Z}  = (\gls{I} - \gls{H}^+\gls{H})$, in which $\gls{I}$ and $\gls{H}$ are the identity and smoothness matrix respectively. The parameter $\gls{forget}_1 > 0$ indicates the confidence in the initial estimated parameters, where a larger $\gls{forget}_1$ indicates a lower confidence level.
To initialize $\gls{Ch}$, it is important to pick these such that the continuity constraints are not violated. For this, a constrained \gls{LS} fit of the estimated shape can be used, using the approach from \citea{deVisser2009a}. If no knowledge is available, initialization of all coefficients at zero will satisfy the constraints, resulting in:
\begin{equation} \gls{Ch}_1 = \mathbf{0} \end{equation}

\begin{table}[t]
\renewcommand{\arraystretch}{1.3}
\caption{\acrshort{RLS} algorithm from \citea{deVisser2011}}
\label{tab:trainMIL}
\centering
 \begin{tabular}{lrll}
  \hline
  \hline
		\textbf{Step} & & \textbf{Action} & \textbf{Computational} \\
		 & & & \textbf{Complexity} \\
    \hline
    (1) & $\gls{e}_{t} =$ & $\gls{y}_{t} - \gls{xb}_{t}^{\top} \gls{Ch}_{t}$ 	& $\mathcal{O}(\gls{dh})$	\\
    (2) & $\gls{P}_{t+1} =$ & $\gls{P}_{t} - \frac{ \gls{P}_{t} \gls{xb}_{t} \gls{xb}_{t}^{\top}  \gls{P}_{t}}{1 + \gls{xb}_{t}^{\top} \gls{P}_{t}  \gls{xb}_{t}} $	& $\mathcal{O}(\gls{ah}^2)$ \\
    (3) & $\gls{Ch}_{t+1} =$ & $\gls{Ch}_{t} + \gls{P}_{t+1} \gls{xb}_{t} \gls{e}_{t}$ & $\mathcal{O}(\gls{ah}^2)$	\\
  \hline
  \hline
 \end{tabular}
\end{table}


\section{Spline Dynamic Programming} \label{sec:sdp}
This section has a dual purpose; it will introduce the framework that combines both simplex splines and the \gls{RLSTD} algorithm, and the modified \gls{RLSTD} algorithm with the ability to track time-varying systems will be defined.

\subsection{The SDP framework} \label{sec:sdpframework}
To successfully represent the optimal value function with a simplex spline, a spline space has to have sufficient approximation power at each point of the value function domain. While it is theoretically possible to have an infinite refinement in terms of triangulation, the idea behind the parametrization of the value function is that there is no need for an infinite amount of states, but only a limited amount of parameters to describe the entire state space $\gls{X}$. 
With no a-priori information available, a triangulation consisting of nodes positioned in a grid is a good initial estimate, as it evenly distributes the approximation power over the domain according to Eq.~\ref{eq:approx}. 
What remains is to construct a spline space $S_d^r (\mathcal{T})$ are the polynomial degree and continuity order, parameters with a global effect on the spline function. Finally, to start the procedure, only the initial coefficients and covariance matrix have to be constructed. 
There are additional settings required for exploration (included in the policy) and the adaptability (to be discussed in the next section).
Note that there is an efficient method to derive the directional derivatives, which are used in the optimal policy, available in \citea{deVisser2011}.

While the framework functions in an infinite-time setting, the simulation has a time limit after which a new trial is started. This to have a successful convergence to the optimal value function by visiting every state due to a random initial state $\gls{x}_0$. At each time step an action is selected, the next state is determined, the reward is calculated and the \gls{RLSTD} algorithm updates the parameter coefficients $\gls{Ch}$ (Eq. \ref{eq:defc}) and covariance matrix $\gls{P}$ (Eq. \ref{eq:p0}). An overview of the complete algorithm is available in Table~\ref{tab:SDP}.

The requirement of each state $\gls{x} \in \gls{X}$ being visited is essential to guarantee the convergence to the optimal coefficients $\gls{Ch}^*$. Therefore an explorer is introduced into the framework that explores the entire state-space $\gls{X}$. Note that this proof assumes that the entire state-space is reachable, which is not true for every environment.

In this situation, the \gls{SDP} framework consists of a direct implementation of the \gls{RLSTD} algorithm, using multivariate simplex B-splines as a linear-in-the-parameters function approximator. As a consequence, the convergence proof as given in \citea{BB96} applies.

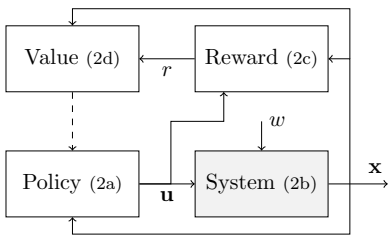
\begin{figure}
  \tikzstyle{block} = [draw, rectangle, fill=gray!10,
      minimum height=3em, minimum width=6em]
  \tikzstyle{block1} = [draw, rectangle, 
      minimum height=3em, minimum width=6em]
  \centering
\resizebox{!}{3cm} {
  \begin{tikzpicture}[auto, node distance=2cm]
      \node [block1] (controller) {Policy \footnotesize{(2a)}};
      \node [block, right of=controller, node distance=3cm] (system) {System \footnotesize{(2b)}};
      \node [coordinate, above of=system, node distance=1cm] (w) {};
      \node [right of=w, node distance=0.25cm] (w2) {$w$};
      \draw [->] (w) -- (system);
      \node [block1, above of=controller] (value) {Value \footnotesize{(2d)}};
      \node [block1, right of=value, node distance=3cm] (reward) {Reward \footnotesize{(2c)}};
      \draw [->] (controller) -- node[name=u, below] {\gls{u}} (system);
      \draw [->] (reward) -- node[name=r] {\gls{r}} (value);
      \node [coordinate, right of=system] (output) {};
      \node [above of=output, node distance=0.25cm, xshift=-0.2cm] (o2) {\gls{x}};
      \node [coordinate, right of=system, node distance = 1.4cm] (y1) {};
      \node [coordinate, right of=reward, node distance = 1.4cm] (y2) {};
      \node [coordinate, node distance = 1cm, above of=r] (top) {};
      \node [coordinate, node distance = 0.6cm, below of=u] (bottom) {};
      \node [coordinate, node distance = 1cm, above of=u] (mid) {};
      \draw [->] (controller.east) -- ([xshift=0.5cm]controller.east) -- ([xshift=0.5cm, yshift=1cm]controller.east) -| ([xshift=-0.6cm]reward.south);
      \draw [->] (system) -- (output);
      \draw [->] (y1) |- (bottom) -| (controller);
      \draw [->] (y1) -- (y2) -- (reward.east);
      \draw [->] (y2) |- (top) -| (value.north);
      \draw [->, dashed] (value) -- (controller);
  \end{tikzpicture}
} 
\caption{The control diagram of the \acrshort{SDP} framework, including the corresponding step from Table~\ref{tab:SDP}}
\label{fig:SDPblock}
\end{figure}

A block diagram of the control scheme is presented in Figure~\ref{fig:SDPblock}. Both the state $\gls{x}$ and input $\gls{u}$ are in the diagram, as well as the reward $\gls{r}$ and the system disturbance $w$. The \gls{DP} components are the Policy, Reward and Value, however where both the Policy and Reward components are identical for \gls{NDP} and \gls{SDP}, the Value component differs. The method to construct the value function in the \gls{SDP} framework is described in section~\ref{sec:policyeval}.

\begin{table}
\caption{The \acrshort{SDP} overview}
\label{tab:SDP}
\centering
 \begin{tabular}{ll}
\hline
\hline
\textbf{Step} & \textbf{Action} \\
\hline
(0) & Initialization \\
 ~ (0a)& Create a spline space $S_d^r (\mathcal{T})$ \\
 ~ (0b)& Set the parameter values $\gls{Ch}_1 = \mathbf{0}$ \\
 ~ (0c)& Set the covariance matrix $\gls{P}_1 = \gls{forget}_1 \cdot (\gls{I} - \gls{H}^+\gls{H})$ \\
(1) & Do for $n = 1, \ldots ,$ trials, \\				
 ~ (1a)& Set the initial state $\gls{x}_1$ \\
 ~ (2) & Do for $t = 1, \ldots ,$ end, \\				
 ~  ~ (2a)& $\gls{u}_t = f(\gls{x}_t,\gls{Ch}_t)$ \\
 ~  ~ (2b)& $\gls{x}_{t+1} = f(\gls{x}_t,\gls{u}_t)$ \\
 ~  ~ (2c)& $\gls{r}_{t+1} = f(\gls{x}_{t+1},\gls{u}_t)$ \\
 ~  ~ (2d)& Update $\gls{Ch}$ and $\gls{P}$ with the \acrshort{RLSTD} algorithm \\
\hline
\hline
 \end{tabular}
\end{table}

\subsection{Recursive weighted least squares}
While the objective function in Eq.~\ref{eq:classicLS} will converge to the classic \gls{LS} solution, it is unable to cope with time-varying systems as it weighs each measurement equally, driving the \gls{P} matrix to zero. For the \gls{RLS} algorithm to track time-varying systems, a popular and effective solution in adaptive control is using the forget factor, changing the quadratic objective function to:
\begin{equation} \gls{COST}_t = \frac{1}{t} \sum_{k=1}^{t} \gls{forget}^{t-k} \left[ e_k \right]^2 \label{eq:weightedLS} \end{equation}
where $\gls{forget}$ represents the forget factor. This equation can be rewritten as:
\begin{equation} \gls{COST}_t = \frac{1}{t} \left[ \gls{forget} \gls{COST}_{t-1} + \left[ e_t \right]^2 \right] \end{equation}
making it clear that that \gls{forget} has a discounting effect on the past errors, reducing the importance given to old data. Therefore, by applying a forget factor the \gls{LS} solution is converted to a \gls{WLS} solution, where the newest data-points have the most influence on parameters. 
According to \citea{WZ91} the forget factor can be applied to the covariance matrix as:
\begin{equation} \gls{P}_{t+1} = \gls{forget}^{-1} ~ \gls{P}_{t} \label{eq:globalforget} \end{equation}
Applying the forget factor in this form has the disadvantage that it scales all elements of \gls{P} equally. This will result in covariance wind-up when no new information is available over a long period, caused by certain elements of \gls{P} becoming very large.
This is a direct consequence of the spatial influence of the B-coefficients, visible in the B-net seen in Figure~\ref{fig:bnet}.
The forget method perceived in Eq.~\ref{eq:globalforget} is therefore only used at initialization, represented by $\gls{forget}_1$ seen in Eq.~\ref{eq:p0}.

A solution to prevent the covariance wind-up is to apply the forget factor only to the updated parameters. This approach is called directional forgetting \citea{WZ91} and updates the covariance matrix as follows:
\begin{equation} \gls{P}_{t+1} = \gls{P}_{t} + \gls{forget}_2 ~ \gls{xb}_{t} \gls{xb}_{t}^{\top} \end{equation}
where $\gls{forget}_{2}$ represents a forget factor applied to the updated parameters.
However, the problem with this approach is that it destroys the continuity by ignoring the constraints set by \gls{H} in Eq.~\ref{eq:matrixH}.
Therefore, in order to keep $\gls{Ch}$ in the null-space of $\gls{H}$, the following update is proposed:
\begin{equation} \gls{P}_{t+1} = \gls{P}_{t} + \gls{forget}_{2} \left[ \mathbf{Z} \gls{xb}_{t} \gls{xb}_{t}^{\top}  \mathbf{Z} \right] \end{equation}
where $\mathbf{Z}  = (\gls{I} - \gls{H}^+\gls{H})$, which is the projection on the null-space of $\mathbf{H}$, introduced in Eq.~\ref{eq:p0}.

This approach can be implemented by altering step (2) in the original \gls{RLS} algorithm from Table~\ref{tab:trainMIL} to:
\begin{equation}
\gls{P}_{t+1} = \gls{P}_{t} - \frac{ \gls{P}_{t} \gls{xb}_{t} \gls{xb}_{t}^{\top}  \gls{P}_{t}}{1 + \gls{xb}_{t}^{\top} \gls{P}_{t}  \gls{xb}_{t}} + \gls{forget}_{2} \left[ \mathbf{Z} \gls{xb}_{t} \gls{xb}_{t}^{\top}  \mathbf{Z} \right] \label{eq:modcov} \end{equation}
resulting in our new formulation for the covariance matrix update step.
The modified \gls{RLSTD} algorithm with the additional term in step (2) is visible in Table~\ref{tab:trainRLSTDmod}, including the computational complexity. To immediately apply the forget factor at time $t$, step (3) employs the $\gls{P}_{t+1}$ matrix, as done in \citea{ljung1983theory}.

The \gls{RLS} algorithm with directional forgetting is simply convergent for a system where the data generation mechanism is deterministic \citea{bittanti1990}. It should be noted that under this assumption, \gls{LMS} algorithms also have proven convergence \citea{TVR97}.

Additionally, the modified \gls{RLSTD} algorithm is capable of filtering out the residual noise to end up near the optimal coefficients $\gls{Ch}^*$. With $\gls{forget}_{2} = 0$, the filter has an infinite window in time, while at $\gls{forget}_{2} > 0$, the window is infinite no longer, which has the advantage of being able to track time-varying systems and disadvantage of being susceptible to noise. This trade-off between noise filtering and tracking is an often returning phenomenon in adaptive control \citea{WZ91}. In principle $\gls{forget}_{2} > 0$ only has a beneficial effect on the control of a time-varying system.
Luckily, this approach allows the use of a variable $\gls{forget}_{2}$, able to increase and decrease as desired. The design of a successful variable forget factor will result in both good tracking behavior and a good performance with residual noise.

\begin{table}
\caption{\acrshort{RLSTD} algorithm, modified from Table~\ref{tab:trainRLSTD}}
\label{tab:trainRLSTDmod}
\centering
 \begin{tabular}{lrll}
  \hline
  \hline
	  \textbf{Step} & & \textbf{Action} & \textbf{Computational} \\
			& & & \textbf{Complexity} 		\\
    \hline
(1)& $e_t =$ & $ \gls{r}_{t+1} - (\gls{xb}_t - \gls{gamma} \gls{xb}_{t+1})^{\top} \gls{Ch}_{t}$ & $\mathcal{O}(\gls{dh})$ \\
(2)& $\gls{P}_{t+1} =$ & $\gls{P}_{t} - \frac{ \gls{P}_{t} \gls{xb}_{t} ( \gls{xb}_t - \gls{gamma} \gls{xb}_{t+1} )^{\top}  \gls{P}_{t}}{1 + ( \gls{xb}_t - \gls{gamma} \gls{xb}_{t+1} )^{\top} \gls{P}_{t} \gls{xb}_t} $ & $\mathcal{O}(\gls{ah}^2)$ \\
   & & $+ \gls{forget}_{2} \left[ \mathbf{Z} \gls{xb}_{t} \gls{xb}_{t}^{\top} \mathbf{Z} \right]$ &  \\
(3) & $ \gls{Ch}_{t+1} =$ & $ \gls{Ch}_{t} +  \gls{P}_{t+1} \gls{xb}_t e_t$ & $\mathcal{O}(\gls{ah}^2)$ \\
  \hline
  \hline
 \end{tabular}
\end{table}

\section{Performance Evaluation of SDP} \label{sec:results}
The proposed \gls{SDP} framework has been implemented on a pendulum swing-up non-linear control problem, as seen in \ref{sec:swingup}.
At the start, the controller has no knowledge about the optimal value function, and has to learn from online measurements. 
The gain input of the plant is assumed to be known; it will increase the learning time for both algorithms with the same amount if it is to be identified using model identification.
The objective of the experiment is to move and keep the pendulum in an upwards position by using a limited torque. 
The controller receives reinforcement at each state, where the top position is most beneficial.
The system dynamics are simulated using an Euler integration scheme in combination with the equations of motion as presented in \ref{sec:swingup}. 

The performance of each trial is measured by the maximum amount of time $t_{up}$ the pendulum is consecutively kept in an upwards position, where the upwards position is defined as:
\begin{equation} | \gls{theta}_{up} | < \frac{\pi}{4} = 45^{\circ} \label{eq:up} \end{equation}
A trial itself consists of $20$ s, with time steps of $0.02$ s. As each trial is initialized in a random angle $\gls{theta}$ and a zero angle rate $\gls{thetad}$ (consistent with the experiment in \citea{MD05}), some trials require more swings to reach the top. Therefore a lower $t_{up}$ does not mean a worse performance per se, but it may have been initialized in a lower position. 
However, while $\gls{theta}_0$ is random, it is identical for each trial over the different methods. This is done to remove the chance of one method having better initializations than the other, degrading the quality of the comparison.

The neural networks used for \gls{NDP} are constructed using either radial basis function or using a sigmoid function as a basis. This corresponds to the radial basis network and feedforward network respectively. In case of the radial basis network, the TD($\lambda$) algorithm is used for training to increase the performance, while the feedforward network is trained using the gradient descent approach. More information on how these networks are constructed and trained is available in \citea{bertsekas1996neuro} and \citea{Roj96}.

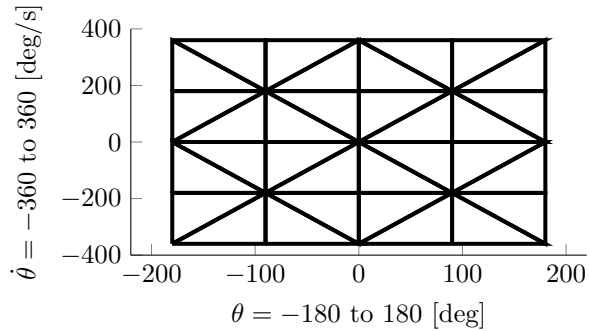
\begin{figure}
\centering
%
%
%
\begin{tikzpicture}

\begin{axis}[%
width=6cm,
height=3cm,
scale only axis,
xmin=-220,
xmax=220,
xlabel={$\theta = -180$ to $180$ [deg]},
ymin=-400,
ymax=400,
ylabel={$\dot \theta = -360$ to $360$ [deg/s]},
axis x line*=bottom,
axis y line*=left
]
\addplot [
color=black,
solid,
line width=1.5pt,
forget plot
]
table[row sep=crcr]{
-180 -360\\
-90 -360\\
-90 -180\\
-180 -360\\
};
\addplot [
color=black,
solid,
line width=1.5pt,
forget plot
]
table[row sep=crcr]{
-180 -360\\
-180 -180\\
-90 -180\\
-180 -360\\
};
\addplot [
color=black,
solid,
line width=1.5pt,
forget plot
]
table[row sep=crcr]{
-180 -180\\
-180 0\\
-90 -180\\
-180 -180\\
};
\addplot [
color=black,
solid,
line width=1.5pt,
forget plot
]
table[row sep=crcr]{
-180 0\\
-90 -180\\
-90 0\\
-180 0\\
};
\addplot [
color=black,
solid,
line width=1.5pt,
forget plot
]
table[row sep=crcr]{
-90 -360\\
-90 -180\\
0 -360\\
-90 -360\\
};
\addplot [
color=black,
solid,
line width=1.5pt,
forget plot
]
table[row sep=crcr]{
-90 -180\\
0 -360\\
0 -180\\
-90 -180\\
};
\addplot [
color=black,
solid,
line width=1.5pt,
forget plot
]
table[row sep=crcr]{
-90 -180\\
0 -180\\
0 0\\
-90 -180\\
};
\addplot [
color=black,
solid,
line width=1.5pt,
forget plot
]
table[row sep=crcr]{
-90 -180\\
-90 0\\
0 0\\
-90 -180\\
};
\addplot [
color=black,
solid,
line width=1.5pt,
forget plot
]
table[row sep=crcr]{
0 -360\\
90 -360\\
90 -180\\
0 -360\\
};
\addplot [
color=black,
solid,
line width=1.5pt,
forget plot
]
table[row sep=crcr]{
0 -360\\
0 -180\\
90 -180\\
0 -360\\
};
\addplot [
color=black,
solid,
line width=1.5pt,
forget plot
]
table[row sep=crcr]{
0 -180\\
0 0\\
90 -180\\
0 -180\\
};
\addplot [
color=black,
solid,
line width=1.5pt,
forget plot
]
table[row sep=crcr]{
0 0\\
90 -180\\
90 0\\
0 0\\
};
\addplot [
color=black,
solid,
line width=1.5pt,
forget plot
]
table[row sep=crcr]{
90 -360\\
90 -180\\
180 -360\\
90 -360\\
};
\addplot [
color=black,
solid,
line width=1.5pt,
forget plot
]
table[row sep=crcr]{
90 -180\\
180 -360\\
180 -180\\
90 -180\\
};
\addplot [
color=black,
solid,
line width=1.5pt,
forget plot
]
table[row sep=crcr]{
90 -180\\
180 -180\\
180 0\\
90 -180\\
};
\addplot [
color=black,
solid,
line width=1.5pt,
forget plot
]
table[row sep=crcr]{
90 -180\\
90 0\\
180 0\\
90 -180\\
};
\addplot [
color=black,
solid,
line width=1.5pt,
forget plot
]
table[row sep=crcr]{
-180 0\\
-90 0\\
-90 180\\
-180 0\\
};
\addplot [
color=black,
solid,
line width=1.5pt,
forget plot
]
table[row sep=crcr]{
-180 0\\
-180 180\\
-90 180\\
-180 0\\
};
\addplot [
color=black,
solid,
line width=1.5pt,
forget plot
]
table[row sep=crcr]{
-180 180\\
-180 360\\
-90 180\\
-180 180\\
};
\addplot [
color=black,
solid,
line width=1.5pt,
forget plot
]
table[row sep=crcr]{
-180 360\\
-90 180\\
-90 360\\
-180 360\\
};
\addplot [
color=black,
solid,
line width=1.5pt,
forget plot
]
table[row sep=crcr]{
-90 0\\
0 0\\
-90 180\\
-90 0\\
};
\addplot [
color=black,
solid,
line width=1.5pt,
forget plot
]
table[row sep=crcr]{
0 0\\
-90 180\\
0 180\\
0 0\\
};
\addplot [
color=black,
solid,
line width=1.5pt,
forget plot
]
table[row sep=crcr]{
-90 180\\
0 180\\
0 360\\
-90 180\\
};
\addplot [
color=black,
solid,
line width=1.5pt,
forget plot
]
table[row sep=crcr]{
-90 180\\
-90 360\\
0 360\\
-90 180\\
};
\addplot [
color=black,
solid,
line width=1.5pt,
forget plot
]
table[row sep=crcr]{
0 0\\
90 0\\
90 180\\
0 0\\
};
\addplot [
color=black,
solid,
line width=1.5pt,
forget plot
]
table[row sep=crcr]{
0 0\\
0 180\\
90 180\\
0 0\\
};
\addplot [
color=black,
solid,
line width=1.5pt,
forget plot
]
table[row sep=crcr]{
0 180\\
0 360\\
90 180\\
0 180\\
};
\addplot [
color=black,
solid,
line width=1.5pt,
forget plot
]
table[row sep=crcr]{
0 360\\
90 180\\
90 360\\
0 360\\
};
\addplot [
color=black,
solid,
line width=1.5pt,
forget plot
]
table[row sep=crcr]{
90 0\\
180 0\\
90 180\\
90 0\\
};
\addplot [
color=black,
solid,
line width=1.5pt,
forget plot
]
table[row sep=crcr]{
180 0\\
90 180\\
180 180\\
180 0\\
};
\addplot [
color=black,
solid,
line width=1.5pt,
forget plot
]
table[row sep=crcr]{
90 180\\
180 180\\
180 360\\
90 180\\
};
\addplot [
color=black,
solid,
line width=1.5pt,
forget plot
]
table[row sep=crcr]{
90 180\\
90 360\\
180 360\\
90 180\\
};
\end{axis}
\end{tikzpicture}%
\caption{Type III Delaunay Triangulation $\mathcal{T}_{32}$}
\label{fig:triangulation}
\end{figure}

For \gls{SDP}, a $4^{th}$ degree spline space with $1^{st}$ order continuity, without ($\gls{forget}_2 = 0$) and with ($\gls{forget}_2 = 0.4$) forget factor (see Eq.~\ref{eq:modcov}) has been selected. The polynomial degree has been determined by trial and error, such that the simplex spline is capable of estimating the optimal value function. 
Because the optimal policy is based on the first derivative and \gls{u} has no rate restrictions, a discontinuous first derivative would give an unfair advantage to \gls{SDP}. Therefore the continuity degree has been chosen such that the first derivative is continuous, identical to \gls{NDP}.
The value of $\gls{forget}_2$ is selected such that an increase of tracking behavior is witnessed in the experiments.
A type III Delaunay triangulation of nodes positioned in a grid is used to produce the triangulation $\mathcal{T}_{32}$ seen in Figure~\ref{fig:triangulation}. As explained in section~\ref{sec:sdpframework}, it is essential that this triangulation is capable of approximating the optimal value function.

The parameters of \gls{NDP} used in the simulation have been selected such that there is a comparable amount of parameters in each function approximator;
the $S_4^1(\mathcal{T}_{32})$ has a total of $480$ coefficients, the feedforward network has $480$ weights, and the 12x12 radial basis network has $432$ weights. The centers of the radial basis network are positioned in a grid, including one centered in $\gls{x} = [0 ~ 0]^{\top}$.
A result of the continuity constraints is that the number of free parameters is less than the total amount of parameters, thus effectively lowering the approximation power. In this case, there are $151$ free parameters as the rank of \gls{H} is $329$.

A search for the best set of learning parameters for the feedforward and radial basis network was performed in an attempt to have a strong comparison between \gls{NDP} and \gls{SDP}. An overview of the parameters used in the experiments is visible in Table~\ref{tab:sumparam}.
The initialization of network weights, center weights or nodes has a significant impact; for the feedforward network and simplex splines half of the initializations failed when the networks weights or nodes were selected randomly, while for the radial basis network only 4\% failed. In this case, success is described as scoring a $t_{up} > 10$ s for at least one trial. In the experiment the center weights and nodes were defined a-priori, removing the dependency on the initialization.

First, in section~\ref{sec:exp1} the four control methods are tasked with controlling a stochastic system, which will demonstrate the influence of system noise. Secondly, in section~\ref{sec:exp2} the methods are tasked with controlling a time-varying system; this is meant to exhibit the adaptability of the control methods.

\begin{table}
\caption{The Learning Parameters of the \gls{NDP} and \gls{SDP} Methods used to obtain the Results}
\label{tab:sumparam}
\centering
  \begin{tabular}{c}
    \acrshort{NDP} - Feedforward \\
    \begin{tabular}{c|c|c|c|c}
	\hline
	\hline
	  \textbf{Parameter}	& $\eta_1$ 	& $\eta_2$ 	& Neurons & Total parameters	\\
	  \hline
	  \textbf{Value}	& $10^{-3}$ 	& $10^{-3}$	& $160$ & $480$ \\
	\hline
	\hline
    \end{tabular}
    \\ \\ \acrshort{NDP} - Radial basis \\
    \begin{tabular}{c|c|c|c|c|c}
	\hline
	\hline
	  \textbf{Parameter}	& $\eta_1$ 	& $\eta_2$ 	& $\lambda$ 	& Neurons & Total parameters	\\
	  \hline
	  \textbf{Value}	& $10^{-3}$ 	& $10^{-2}$ 	& $0.8$ 		& 12x12 & $432$	\\
	\hline
	\hline
    \end{tabular}
    \\ \\ \acrshort{SDP}  \\
    \begin{tabular}{c|c|c|c|c|c}
	\hline
	\hline
	  \textbf{Parameter} & $S_d^r$ & $\gls{forget}_1$ & $\gls{forget}_2$ & Triangulation & Total parameters \\
	  \hline
	  \textbf{Value} & $S_4^1$ &  $10$ & $0$ / $0.4$ & Type III $\mathcal{T}_{32}$ & $480$	\\
	\hline
	\hline
    \end{tabular}
  \end{tabular}
\end{table}

\subsection{Experiment I - The Stochastic System} \label{sec:exp1}
In the first experiment the four methods are tasked with controlling a stochastic system. 
The stochastic system has a system disturbance of $\sigma_w = 3^{\circ}/s^2$. This means that the standard deviation of the system noise is $60\%$ of the input's influence, as the maximum influence is $T^{\max}/ml^2 = 5^{\circ}/s^2$.
Since the system noise is applied to the equation of motion, the effect of the noise is only directly connected to $\gls{thetadd}$.
Furthermore, in order to identify the influence of system noise on each method, the deterministic system ($\sigma_w = 0^{\circ}/s^2$) is used as a baseline.

The results of the four methods are visible in the figures~\ref{fig:e1_ndp_w0} and \ref{fig:e1_sdp_w0} for the deterministic system, and in the figures~\ref{fig:e1_ndp_w3} and \ref{fig:e1_sdp_w3} for the stochastic system.
It shows that while the learning parameters and initialized weights are identical for both systems, the \gls{NDP} radial basis has learned to swing the pendulum up in less trials for the stochastic system. This increased learning rate of a dynamic programming algorithm in a stochastic system is a well known phenomenon, and is a result of the extra exploration that occurs due to the system noise \citea{bertsekas1996neuro}. 
Nevertheless the learning rate of \gls{SDP} remains the highest in both the deterministic and stochastic system.

Another observation is that the stability of \gls{NDP} is effected most by the presence of system noise, visible by an overall decrease of $t_{up}$. For \gls{SDP}, only \gls{SDP} - $\gls{forget}_2 = 0.4$ shows a decrease of $t_{up}$ larger than $5$ s in trial $36$ and $68$, thus it can be concluded that \gls{SDP} shows most resilience to system noise.
To support this claim with numerical arguments both the mean and standard deviation of $t_{up}$ of the $100$ trials are presented in Table~\ref{tab:results1}.
While it does not reveal the decrease in stability of the \gls{NDP} methods because it is masked by the increased learning rate, it does show that \gls{SDP} outperforms \gls{NDP} in both scenarios.

\begin{table}
\caption{Mean and standard deviation (std) of $t_{up}$ of the $100$ trials of Experiment I}
\label{tab:results1}
\centering
\begin{tabular}{c}
      $\sigma_w = 0^{\circ}/s^2$ \\
    \begin{tabular}{c|c|c|c|c}    
	\hline
	\hline
			& \acrshort{NDP} - 	& \acrshort{NDP} - 	& \acrshort{SDP} -  	& \acrshort{SDP} -  \\
			& FF 		& RBF 		& $\gls{forget}_{2} = 0$ & $\gls{forget}_{2} = 0.4$ \\
	  \hline
	  \textbf{Mean}	& $12.71$ s 	& $13.60$ s	& $18.30$ s	& $18.16$ s	\\
	  \textbf{Std}	& $7.89$ s	& $7.80$ s	& $2.84$ s	& $3.24$ s	\\
	\hline
	\hline
    \end{tabular} \\ \\
      $\sigma_w = 3^{\circ}/s^2$ \\
    \begin{tabular}{c|c|c|c|c}    
	\hline
	\hline
			& \acrshort{NDP} - 	& \acrshort{NDP} - 	& \acrshort{SDP} -  	& \acrshort{SDP} -  \\
			& FF 		& RBF 		& $\gls{forget}_{2} = 0$ & $\gls{forget}_{2} = 0.4$ \\
	  \hline
	  \textbf{Mean}	& $11.95$ s	& $15.48$ s	& $18.14$ s	& $17.86$ s	\\
	  \textbf{Std}	& $6.62$ s	& $4.93$ s	& $3.08$ s	& $2.91$ s	\\
	\hline
	\hline
    \end{tabular}
\end{tabular}
\end{table}


\newcommand{\tikzwidth}{7cm}
\newcommand{\tikzheight}{4cm}

\begin{figure*}
	\centering
%
%
%
\begin{tikzpicture}

\begin{axis}[%
width=\tikzwidth,
height=\tikzheight,
scale only axis,
xmin=0,
xmax=99,
xlabel={Trial},
xmajorgrids,
ymin=-0.1,
ymax=20.1,
ylabel={$t_{up}$ [s]},
ymajorgrids,
axis x line*=bottom,
axis y line*=left,
legend style={at={(1.03,0.5)},anchor=west,draw=black,fill=white,legend cell align=left}
]
\addplot [
color=blue,
line width=0.3pt,
mark size=1.5pt,
only marks,
mark=o,
mark options={solid}
]
table[row sep=crcr]{
1 2.62\\
2 3.7\\
3 0.5\\
4 0.52\\
5 0.94\\
6 0.46\\
7 1.04\\
8 0.42\\
9 2.18\\
10 0.58\\
11 0.5\\
12 2.34\\
13 3.84\\
14 4.3\\
15 1.22\\
16 0.88\\
17 0.74\\
18 0.56\\
19 0.58\\
20 0.8\\
21 0.64\\
22 0.74\\
23 0.88\\
24 0.78\\
25 0.84\\
26 0.84\\
27 0.82\\
28 0.82\\
29 0.92\\
30 9\\
31 0.96\\
32 0.96\\
33 15.68\\
34 18.9\\
35 18.8\\
36 19.98\\
37 18.48\\
38 18.64\\
39 19.98\\
40 18.82\\
41 16.26\\
42 18.44\\
43 17.3\\
44 18.54\\
45 17.72\\
46 16.08\\
47 14.76\\
48 16.38\\
49 18.78\\
50 17.16\\
51 19.98\\
52 18.82\\
53 16.5\\
54 16.06\\
55 16.5\\
56 16.46\\
57 17.44\\
58 18.64\\
59 17.72\\
60 18.56\\
61 16.6\\
62 19.98\\
63 16.74\\
64 19.98\\
65 17.46\\
66 17.9\\
67 18.56\\
68 18.88\\
69 17.42\\
70 18.66\\
71 16.66\\
72 17.62\\
73 16.02\\
74 18.68\\
75 17.52\\
76 19.98\\
77 15.98\\
78 19.98\\
79 16.92\\
80 16.12\\
81 17\\
82 17.56\\
83 18.4\\
84 19.98\\
85 18.82\\
86 18.78\\
87 16.04\\
88 18.88\\
89 19.98\\
90 18.76\\
91 17.56\\
92 18.72\\
93 18.76\\
94 16.58\\
95 19.98\\
96 19.98\\
97 18.58\\
98 19.98\\
99 17.14\\
};
\addlegendentry{NDP - FF};

\addplot [
color=blue,
dash pattern=on 1pt off 3pt on 3pt off 3pt,
line width=0.8pt
]
table[row sep=crcr]{
3 1.656\\
4 1.224\\
5 0.692\\
6 0.676\\
7 1.008\\
8 0.936\\
9 0.944\\
10 1.204\\
11 1.888\\
12 2.312\\
13 2.44\\
14 2.516\\
15 2.196\\
16 1.54\\
17 0.796\\
18 0.712\\
19 0.664\\
20 0.664\\
21 0.728\\
22 0.768\\
23 0.776\\
24 0.816\\
25 0.832\\
26 0.82\\
27 0.848\\
28 2.48\\
29 2.504\\
30 2.532\\
31 5.504\\
32 9.1\\
33 11.06\\
34 14.864\\
35 18.368\\
36 18.96\\
37 19.176\\
38 19.18\\
39 18.436\\
40 18.428\\
41 18.16\\
42 17.872\\
43 17.652\\
44 17.616\\
45 16.88\\
46 16.696\\
47 16.744\\
48 16.632\\
49 17.412\\
50 18.224\\
51 18.248\\
52 17.704\\
53 17.572\\
54 16.868\\
55 16.592\\
56 17.02\\
57 17.352\\
58 17.764\\
59 17.792\\
60 18.3\\
61 17.92\\
62 18.372\\
63 18.152\\
64 18.412\\
65 18.128\\
66 18.556\\
67 18.044\\
68 18.284\\
69 18.036\\
70 17.848\\
71 17.276\\
72 17.528\\
73 17.3\\
74 17.964\\
75 17.636\\
76 18.428\\
77 18.076\\
78 17.796\\
79 17.2\\
80 17.516\\
81 17.2\\
82 17.812\\
83 18.352\\
84 18.708\\
85 18.404\\
86 18.5\\
87 18.5\\
88 18.488\\
89 18.244\\
90 18.78\\
91 18.756\\
92 18.076\\
93 18.32\\
94 18.804\\
95 18.776\\
96 19.02\\
97 19.132\\
98 18.624\\
};
\addlegendentry{NDP - FF};
\addlegendimage{empty legend};
\addlegendentry{trend};

\addplot [
color=black,
line width=0.3pt,
mark size=1.5pt,
only marks,
mark=asterisk,
mark options={solid}
]
table[row sep=crcr]{
1 1.2\\
2 0\\
3 0.72\\
4 0\\
5 1\\
6 1.2\\
7 0.78\\
8 0.56\\
9 1.62\\
10 1.9\\
11 1.94\\
12 1.3\\
13 3.46\\
14 2.08\\
15 1.7\\
16 1.58\\
17 0.92\\
18 0.8\\
19 1\\
20 1.1\\
21 1.16\\
22 1.22\\
23 0.94\\
24 1.48\\
25 1.18\\
26 1.4\\
27 0.94\\
28 1.3\\
29 1.22\\
30 1.76\\
31 0\\
32 1.44\\
33 1.4\\
34 1.4\\
35 1.46\\
36 1.7\\
37 1.74\\
38 1.64\\
39 1.62\\
40 1.7\\
41 1.38\\
42 1.52\\
43 1.44\\
44 1.36\\
45 1.76\\
46 1.56\\
47 1.72\\
48 1.68\\
49 1.62\\
50 1.64\\
51 1.62\\
52 1.68\\
53 1.7\\
54 1.52\\
55 1.78\\
56 1.8\\
57 1.58\\
58 1.48\\
59 1.62\\
60 1.36\\
61 1.26\\
62 1.64\\
63 1.66\\
64 2.2\\
65 2.2\\
66 1.7\\
67 1.64\\
68 1.9\\
69 1.86\\
70 1.88\\
71 1.84\\
72 1.82\\
73 1.84\\
74 1.7\\
75 1.7\\
76 2.32\\
77 2.5\\
78 2.86\\
79 3.02\\
80 3.3\\
81 2.78\\
82 2.64\\
83 2.14\\
84 3.88\\
85 3.56\\
86 3.54\\
87 4\\
88 4.36\\
89 5.9\\
90 5.34\\
91 6.28\\
92 7.46\\
93 7.94\\
94 10.54\\
95 19.98\\
96 19.98\\
97 18.88\\
98 19.98\\
99 17.74\\
};
\addlegendentry{NDP - RBF};

\addplot [
color=black,
solid,
line width=0.8pt
]
table[row sep=crcr]{
3 0.584\\
4 0.584\\
5 0.74\\
6 0.708\\
7 1.032\\
8 1.212\\
9 1.36\\
10 1.464\\
11 2.044\\
12 2.136\\
13 2.096\\
14 2.024\\
15 1.948\\
16 1.416\\
17 1.2\\
18 1.08\\
19 0.996\\
20 1.056\\
21 1.084\\
22 1.18\\
23 1.196\\
24 1.244\\
25 1.188\\
26 1.26\\
27 1.208\\
28 1.324\\
29 1.044\\
30 1.144\\
31 1.164\\
32 1.2\\
33 1.14\\
34 1.48\\
35 1.54\\
36 1.588\\
37 1.632\\
38 1.68\\
39 1.616\\
40 1.572\\
41 1.532\\
42 1.48\\
43 1.492\\
44 1.528\\
45 1.568\\
46 1.616\\
47 1.668\\
48 1.644\\
49 1.656\\
50 1.648\\
51 1.652\\
52 1.632\\
53 1.66\\
54 1.696\\
55 1.676\\
56 1.632\\
57 1.652\\
58 1.568\\
59 1.46\\
60 1.472\\
61 1.508\\
62 1.624\\
63 1.792\\
64 1.88\\
65 1.88\\
66 1.928\\
67 1.86\\
68 1.796\\
69 1.824\\
70 1.86\\
71 1.848\\
72 1.816\\
73 1.78\\
74 1.876\\
75 2.012\\
76 2.216\\
77 2.48\\
78 2.8\\
79 2.892\\
80 2.92\\
81 2.776\\
82 2.948\\
83 3\\
84 3.152\\
85 3.424\\
86 3.868\\
87 4.272\\
88 4.628\\
89 5.176\\
90 5.868\\
91 6.584\\
92 7.512\\
93 10.44\\
94 13.18\\
95 15.464\\
96 17.872\\
97 19.312\\
98 18.856\\
};
\addlegendentry{NDP - RBF};
\addlegendimage{empty legend};
\addlegendentry{trend};

\end{axis}
\end{tikzpicture}%
	\caption{Performance overview of \acrshort{NDP} - Radial basis 12x12 ($432$ parameters) and feedforward network 160 neurons ($480$ parameters), without system noise ($\sigma_w = 0^{\circ}/s^2$). The trendline respresents a 5 point moving average.}
	\label{fig:e1_ndp_w0}
\end{figure*}
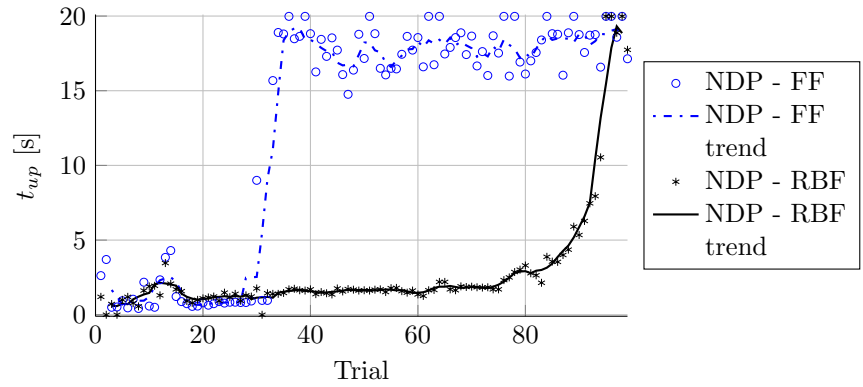

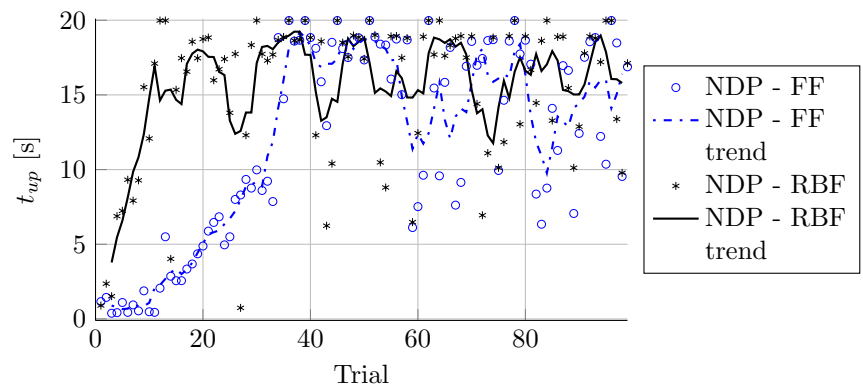
\begin{figure*}
	\centering
%
%
%
\begin{tikzpicture}

\begin{axis}[%
width=\tikzwidth,
height=\tikzheight,
scale only axis,
xmin=0,
xmax=99,
xlabel={Trial},
xmajorgrids,
ymin=-0.1,
ymax=20.1,
ylabel={$t_{up}$ [s]},
ymajorgrids,
axis x line*=bottom,
axis y line*=left,
legend style={at={(1.03,0.5)},anchor=west,draw=black,fill=white,legend cell align=left}
]
\addplot [
color=blue,
line width=0.3pt,
mark size=1.5pt,
only marks,
mark=o,
mark options={solid}
]
table[row sep=crcr]{
1 1.16\\
2 1.44\\
3 0.38\\
4 0.42\\
5 1.1\\
6 0.44\\
7 0.94\\
8 0.54\\
9 1.88\\
10 0.48\\
11 0.44\\
12 2.06\\
13 5.5\\
14 2.86\\
15 2.56\\
16 2.56\\
17 3.34\\
18 3.68\\
19 4.36\\
20 4.88\\
21 5.88\\
22 6.46\\
23 6.84\\
24 4.96\\
25 5.5\\
26 8\\
27 8.3\\
28 9.34\\
29 8.76\\
30 9.98\\
31 8.6\\
32 9.22\\
33 7.86\\
34 18.84\\
35 14.74\\
36 19.98\\
37 18.6\\
38 18.66\\
39 19.98\\
40 18.84\\
41 18.12\\
42 15.88\\
43 12.94\\
44 18.52\\
45 19.98\\
46 18.1\\
47 17.5\\
48 18.64\\
49 18.82\\
50 17.34\\
51 19.98\\
52 18.88\\
53 18.4\\
54 18.34\\
55 16.06\\
56 18.74\\
57 15.02\\
58 18.68\\
59 6.12\\
60 7.52\\
61 9.62\\
62 19.98\\
63 15.46\\
64 9.58\\
65 15.84\\
66 18.18\\
67 7.62\\
68 9.14\\
69 16.92\\
70 18.58\\
71 16.98\\
72 17.44\\
73 18.64\\
74 18.7\\
75 9.94\\
76 14.64\\
77 18.6\\
78 19.98\\
79 17.74\\
80 18.66\\
81 17.06\\
82 8.36\\
83 6.34\\
84 8.76\\
85 14.1\\
86 11.28\\
87 16.96\\
88 16.64\\
89 7.06\\
90 12.42\\
91 17.52\\
92 18.56\\
93 18.84\\
94 12.22\\
95 10.36\\
96 19.98\\
97 18.48\\
98 9.54\\
99 16.88\\
};
\addlegendentry{NDP - FF};

\addplot [
color=blue,
dash pattern=on 1pt off 3pt on 3pt off 3pt,
line width=0.8pt
]
table[row sep=crcr]{
3 0.9\\
4 0.756\\
5 0.656\\
6 0.688\\
7 0.98\\
8 0.856\\
9 0.856\\
10 1.08\\
11 2.072\\
12 2.268\\
13 2.684\\
14 3.108\\
15 3.364\\
16 3\\
17 3.3\\
18 3.764\\
19 4.428\\
20 5.052\\
21 5.684\\
22 5.804\\
23 5.928\\
24 6.352\\
25 6.72\\
26 7.22\\
27 7.98\\
28 8.876\\
29 8.996\\
30 9.18\\
31 8.884\\
32 10.9\\
33 11.852\\
34 14.128\\
35 16.004\\
36 18.164\\
37 18.392\\
38 19.212\\
39 18.84\\
40 18.296\\
41 17.152\\
42 16.86\\
43 17.088\\
44 17.084\\
45 17.408\\
46 18.548\\
47 18.608\\
48 18.08\\
49 18.456\\
50 18.732\\
51 18.684\\
52 18.588\\
53 18.332\\
54 18.084\\
55 17.312\\
56 17.368\\
57 14.924\\
58 13.216\\
59 11.392\\
60 12.384\\
61 11.74\\
62 12.432\\
63 14.096\\
64 15.808\\
65 13.336\\
66 12.072\\
67 13.54\\
68 14.088\\
69 13.848\\
70 15.812\\
71 17.712\\
72 18.068\\
73 16.34\\
74 15.872\\
75 16.104\\
76 16.372\\
77 16.18\\
78 17.924\\
79 18.408\\
80 16.36\\
81 13.632\\
82 11.836\\
83 10.924\\
84 9.768\\
85 11.488\\
86 13.548\\
87 13.208\\
88 12.872\\
89 14.12\\
90 14.44\\
91 14.88\\
92 15.912\\
93 15.5\\
94 15.992\\
95 15.976\\
96 14.116\\
97 15.048\\
98 16.412\\
};
\addlegendentry{NDP - FF};
\addlegendimage{empty legend};
\addlegendentry{trend};

\addplot [
color=black,
line width=0.3pt,
mark size=1.5pt,
only marks,
mark=asterisk,
mark options={solid}
]
table[row sep=crcr]{
1 0.9\\
2 2.36\\
3 1.52\\
4 6.88\\
5 7.24\\
6 9.32\\
7 7.92\\
8 9.28\\
9 15.52\\
10 12.08\\
11 17.1\\
12 19.98\\
13 19.98\\
14 4.02\\
15 15.34\\
16 17.46\\
17 16.56\\
18 18.56\\
19 17.46\\
20 18.74\\
21 18.84\\
22 15.98\\
23 16.72\\
24 17.4\\
25 13.8\\
26 17.74\\
27 0.74\\
28 12.3\\
29 18.34\\
30 19.98\\
31 17.74\\
32 17.3\\
33 17.7\\
34 18.68\\
35 18.88\\
36 19.98\\
37 18.64\\
38 18.7\\
39 19.98\\
40 18.84\\
41 12.3\\
42 18.58\\
43 6.24\\
44 10.4\\
45 19.98\\
46 18.5\\
47 17.56\\
48 18.98\\
49 18.86\\
50 17.42\\
51 19.98\\
52 19\\
53 10.48\\
54 8.8\\
55 18.92\\
56 18.94\\
57 17.48\\
58 18.8\\
59 6.48\\
60 12.44\\
61 18.9\\
62 19.98\\
63 17.7\\
64 19.98\\
65 17.62\\
66 18.32\\
67 18.78\\
68 18.94\\
69 17.5\\
70 18.92\\
71 14.4\\
72 6.94\\
73 11.1\\
74 18.84\\
75 10.12\\
76 11.84\\
77 18.94\\
78 19.98\\
79 13.04\\
80 18.92\\
81 16.68\\
82 14.46\\
83 18.62\\
84 19.98\\
85 13.28\\
86 18.88\\
87 18.9\\
88 15.46\\
89 10.12\\
90 12.88\\
91 17.78\\
92 18.92\\
93 18.8\\
94 17.2\\
95 19.98\\
96 19.98\\
97 13.38\\
98 9.78\\
99 17.12\\
};
\addlegendentry{NDP - RBF};

\addplot [
color=black,
solid,
line width=0.8pt
]
table[row sep=crcr]{
3 3.78\\
4 5.464\\
5 6.576\\
6 8.128\\
7 9.856\\
8 10.824\\
9 12.38\\
10 14.792\\
11 16.932\\
12 14.632\\
13 15.284\\
14 15.356\\
15 14.672\\
16 14.388\\
17 17.076\\
18 17.756\\
19 18.032\\
20 17.916\\
21 17.548\\
22 17.536\\
23 16.548\\
24 16.328\\
25 13.28\\
26 12.396\\
27 12.584\\
28 13.82\\
29 13.82\\
30 17.132\\
31 18.212\\
32 18.28\\
33 18.06\\
34 18.508\\
35 18.776\\
36 18.976\\
37 19.236\\
38 19.228\\
39 17.692\\
40 17.68\\
41 15.188\\
42 13.272\\
43 13.5\\
44 14.74\\
45 14.536\\
46 17.084\\
47 18.776\\
48 18.264\\
49 18.56\\
50 18.848\\
51 17.148\\
52 15.136\\
53 15.436\\
54 15.228\\
55 14.924\\
56 16.588\\
57 16.124\\
58 14.828\\
59 14.82\\
60 15.32\\
61 15.1\\
62 17.8\\
63 18.836\\
64 18.72\\
65 18.48\\
66 18.728\\
67 18.232\\
68 18.492\\
69 17.708\\
70 15.34\\
71 13.772\\
72 14.04\\
73 12.28\\
74 11.768\\
75 14.168\\
76 15.944\\
77 14.784\\
78 16.544\\
79 17.512\\
80 16.616\\
81 16.344\\
82 17.732\\
83 16.604\\
84 17.044\\
85 17.932\\
86 17.3\\
87 15.328\\
88 15.248\\
89 15.028\\
90 15.032\\
91 15.7\\
92 17.116\\
93 18.536\\
94 18.976\\
95 17.868\\
96 16.064\\
97 16.048\\
98 15.796\\
};
\addlegendentry{NDP - RBF};
\addlegendimage{empty legend};
\addlegendentry{trend};

\end{axis}
\end{tikzpicture}%
	\caption{Performance overview of \acrshort{NDP} - Radial basis 12x12 ($432$ parameters) and feedforward network 160 neurons ($480$ parameters), with system noise ($\sigma_w = 3^{\circ}/s^2$). The trendline respresents a 5 point moving average.}
	\label{fig:e1_ndp_w3}
\end{figure*}

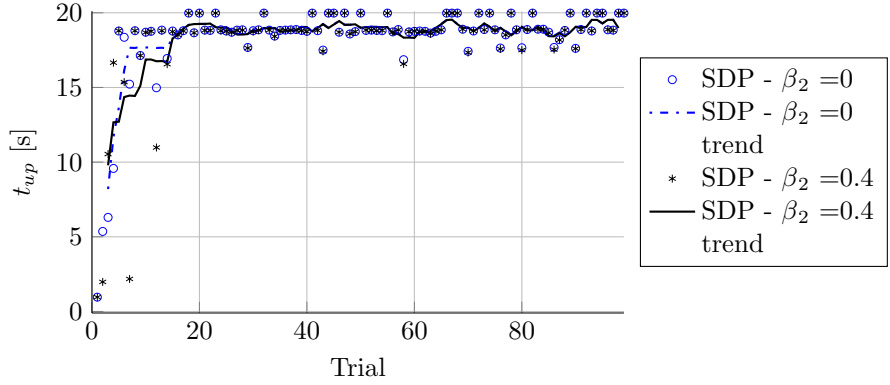
\begin{figure*}
	\centering
	{
%
%
%
\begin{tikzpicture}

\begin{axis}[%
width=\tikzwidth,
height=\tikzheight,
scale only axis,
xmin=0,
xmax=99,
xlabel={Trial},
xmajorgrids,
ymin=-0.1,
ymax=20.1,
ylabel={$t_{up}$ [s]},
ymajorgrids,
axis x line*=bottom,
axis y line*=left,
legend style={at={(1.03,0.5)},anchor=west,draw=black,fill=white,legend cell align=left}
]
\addplot [
color=blue,
line width=0.3pt,
mark size=1.5pt,
only marks,
mark=o,
mark options={solid}
]
table[row sep=crcr]{
1 0.96\\
2 5.36\\
3 6.3\\
4 9.58\\
5 18.78\\
6 18.36\\
7 15.22\\
8 18.8\\
9 17.14\\
10 18.7\\
11 18.76\\
12 14.98\\
13 18.84\\
14 16.94\\
15 18.78\\
16 18.52\\
17 18.8\\
18 19.98\\
19 18.66\\
20 19.98\\
21 18.84\\
22 18.82\\
23 19.98\\
24 18.86\\
25 18.78\\
26 18.7\\
27 18.84\\
28 18.86\\
29 17.68\\
30 18.8\\
31 18.86\\
32 19.98\\
33 18.84\\
34 18.44\\
35 18.82\\
36 18.84\\
37 18.86\\
38 18.88\\
39 18.82\\
40 18.76\\
41 19.98\\
42 18.8\\
43 17.5\\
44 19.98\\
45 19.98\\
46 18.7\\
47 19.98\\
48 18.58\\
49 18.74\\
50 19.98\\
51 18.82\\
52 18.84\\
53 18.82\\
54 18.8\\
55 19.98\\
56 18.72\\
57 18.88\\
58 16.86\\
59 18.72\\
60 18.76\\
61 18.78\\
62 18.76\\
63 18.64\\
64 18.76\\
65 18.86\\
66 19.98\\
67 19.98\\
68 19.98\\
69 18.9\\
70 17.42\\
71 18.78\\
72 19.98\\
73 18.88\\
74 19.98\\
75 18.86\\
76 17.64\\
77 18.52\\
78 19.98\\
79 18.86\\
80 17.66\\
81 18.88\\
82 19.98\\
83 18.9\\
84 18.9\\
85 18.7\\
86 17.68\\
87 18.2\\
88 18.82\\
89 19.98\\
90 17.62\\
91 18.88\\
92 19.98\\
93 18.78\\
94 19.98\\
95 19.98\\
96 18.86\\
97 18.84\\
98 19.98\\
99 19.98\\
};
\addlegendentry{$\text{SDP - }\beta{}_{\text{2}}\text{ =0}$};

\addplot [
color=blue,
dash pattern=on 1pt off 3pt on 3pt off 3pt,
line width=0.8pt
]
table[row sep=crcr]{
3 8.196\\
4 11.676\\
5 13.648\\
6 16.148\\
7 17.66\\
8 17.644\\
9 17.724\\
10 17.676\\
11 17.684\\
12 17.644\\
13 17.66\\
14 17.612\\
15 18.376\\
16 18.604\\
17 18.948\\
18 19.188\\
19 19.252\\
20 19.256\\
21 19.256\\
22 19.296\\
23 19.056\\
24 19.028\\
25 19.032\\
26 18.808\\
27 18.572\\
28 18.576\\
29 18.608\\
30 18.836\\
31 18.832\\
32 18.984\\
33 18.988\\
34 18.984\\
35 18.76\\
36 18.768\\
37 18.844\\
38 18.832\\
39 19.06\\
40 19.048\\
41 18.772\\
42 19.004\\
43 19.248\\
44 18.992\\
45 19.228\\
46 19.444\\
47 19.196\\
48 19.196\\
49 19.22\\
50 18.992\\
51 19.04\\
52 19.052\\
53 19.052\\
54 19.032\\
55 19.04\\
56 18.648\\
57 18.632\\
58 18.388\\
59 18.4\\
60 18.376\\
61 18.732\\
62 18.74\\
63 18.76\\
64 19\\
65 19.244\\
66 19.512\\
67 19.54\\
68 19.252\\
69 19.012\\
70 19.012\\
71 18.792\\
72 19.008\\
73 19.296\\
74 19.068\\
75 18.776\\
76 18.996\\
77 18.772\\
78 18.532\\
79 18.78\\
80 19.072\\
81 18.856\\
82 18.864\\
83 19.072\\
84 18.832\\
85 18.476\\
86 18.46\\
87 18.676\\
88 18.46\\
89 18.7\\
90 19.056\\
91 19.048\\
92 19.048\\
93 19.52\\
94 19.516\\
95 19.288\\
96 19.528\\
97 19.528\\
98 18.996\\
};
\addlegendentry{$\text{SDP - }\beta{}_{\text{2}}\text{ =0}$};
\addlegendimage{empty legend};
\addlegendentry{trend};

\addplot [
color=black,
line width=0.3pt,
mark size=1.5pt,
only marks,
mark=asterisk,
mark options={solid}
]
table[row sep=crcr]{
1 0.96\\
2 1.98\\
3 10.54\\
4 16.66\\
5 18.8\\
6 15.36\\
7 2.18\\
8 18.8\\
9 17.14\\
10 18.68\\
11 18.76\\
12 10.98\\
13 18.8\\
14 16.56\\
15 18.74\\
16 18.52\\
17 18.78\\
18 19.98\\
19 18.66\\
20 19.98\\
21 18.82\\
22 18.82\\
23 19.98\\
24 18.8\\
25 18.76\\
26 18.68\\
27 18.82\\
28 18.84\\
29 17.64\\
30 18.74\\
31 18.84\\
32 19.98\\
33 18.84\\
34 18.42\\
35 18.8\\
36 18.84\\
37 18.84\\
38 18.86\\
39 18.8\\
40 18.74\\
41 19.98\\
42 18.82\\
43 17.42\\
44 19.98\\
45 19.98\\
46 18.7\\
47 19.98\\
48 18.6\\
49 18.74\\
50 19.98\\
51 18.84\\
52 18.84\\
53 18.8\\
54 18.82\\
55 19.98\\
56 18.72\\
57 18.86\\
58 16.58\\
59 18.74\\
60 18.78\\
61 18.74\\
62 18.76\\
63 18.6\\
64 18.76\\
65 18.86\\
66 19.98\\
67 19.98\\
68 19.98\\
69 18.9\\
70 17.34\\
71 18.78\\
72 19.98\\
73 18.86\\
74 19.98\\
75 18.86\\
76 17.58\\
77 18.52\\
78 19.98\\
79 18.86\\
80 17.5\\
81 18.88\\
82 19.98\\
83 18.9\\
84 18.9\\
85 18.66\\
86 17.54\\
87 18.18\\
88 18.84\\
89 19.98\\
90 17.6\\
91 18.88\\
92 19.98\\
93 18.8\\
94 19.98\\
95 19.98\\
96 18.88\\
97 18.82\\
98 19.98\\
99 19.98\\
};
\addlegendentry{$\text{SDP - }\beta{}_{\text{2}}\text{ =0.4}$};

\addplot [
color=black,
solid,
line width=0.8pt
]
table[row sep=crcr]{
3 9.788\\
4 12.668\\
5 12.708\\
6 14.36\\
7 14.456\\
8 14.432\\
9 15.112\\
10 16.872\\
11 16.872\\
12 16.756\\
13 16.768\\
14 16.72\\
15 18.28\\
16 18.516\\
17 18.936\\
18 19.184\\
19 19.244\\
20 19.252\\
21 19.252\\
22 19.28\\
23 19.036\\
24 19.008\\
25 19.008\\
26 18.78\\
27 18.548\\
28 18.544\\
29 18.576\\
30 18.808\\
31 18.808\\
32 18.964\\
33 18.976\\
34 18.976\\
35 18.748\\
36 18.752\\
37 18.828\\
38 18.816\\
39 19.044\\
40 19.04\\
41 18.752\\
42 18.988\\
43 19.236\\
44 18.98\\
45 19.212\\
46 19.448\\
47 19.2\\
48 19.2\\
49 19.228\\
50 19\\
51 19.04\\
52 19.056\\
53 19.056\\
54 19.032\\
55 19.036\\
56 18.592\\
57 18.576\\
58 18.336\\
59 18.34\\
60 18.32\\
61 18.724\\
62 18.728\\
63 18.744\\
64 18.992\\
65 19.236\\
66 19.512\\
67 19.54\\
68 19.236\\
69 18.996\\
70 18.996\\
71 18.772\\
72 18.988\\
73 19.292\\
74 19.052\\
75 18.76\\
76 18.984\\
77 18.76\\
78 18.488\\
79 18.748\\
80 19.04\\
81 18.824\\
82 18.832\\
83 19.064\\
84 18.796\\
85 18.436\\
86 18.424\\
87 18.64\\
88 18.428\\
89 18.696\\
90 19.056\\
91 19.048\\
92 19.048\\
93 19.524\\
94 19.524\\
95 19.292\\
96 19.528\\
97 19.528\\
98 18.98\\
};
\addlegendentry{$\text{SDP - }\beta{}_{\text{2}}\text{ =0.4}$};
\addlegendimage{empty legend};
\addlegendentry{trend};

\end{axis}
\end{tikzpicture}%
}
	\caption{Performance overview of \acrshort{SDP} - $S_4^1 (\mathcal{T}_{32})$ ($480$ parameters), with ($\gls{forget}_{2} = 0.4$) and without ($\gls{forget}_{2} = 0$) forget factor, and without system noise ($\sigma_w = 0^{\circ}/s^2$). The trendline respresents a 5 point moving average.}
	\label{fig:e1_sdp_w0}
\end{figure*}

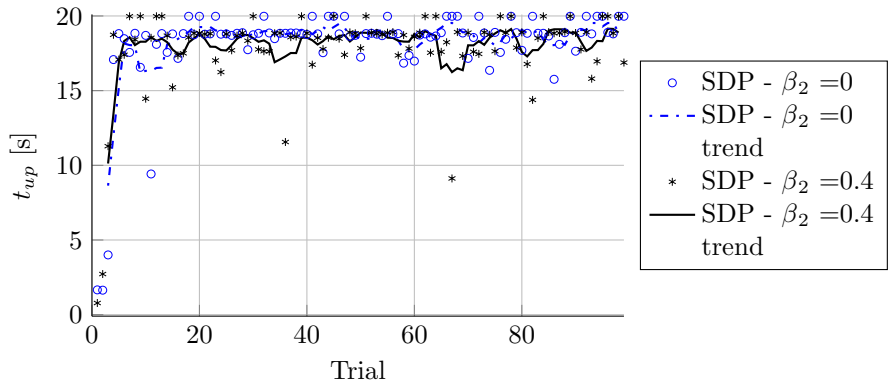
\begin{figure*}
	\centering
	{
%
%
%
\begin{tikzpicture}

\begin{axis}[%
width=\tikzwidth,
height=\tikzheight,
scale only axis,
xmin=0,
xmax=99,
xlabel={Trial},
xmajorgrids,
ymin=-0.1,
ymax=20.1,
ylabel={$t_{up}$ [s]},
ymajorgrids,
axis x line*=bottom,
axis y line*=left,
legend style={at={(1.03,0.5)},anchor=west,draw=black,fill=white,legend cell align=left}
]
\addplot [
color=blue,
line width=0.3pt,
mark size=1.5pt,
only marks,
mark=o,
mark options={solid}
]
table[row sep=crcr]{
1 1.66\\
2 1.64\\
3 4\\
4 17.08\\
5 18.82\\
6 18.48\\
7 17.56\\
8 18.84\\
9 16.56\\
10 18.7\\
11 9.42\\
12 18.1\\
13 18.82\\
14 17.56\\
15 18.8\\
16 17.16\\
17 18.82\\
18 19.98\\
19 18.74\\
20 19.98\\
21 18.78\\
22 18.8\\
23 19.98\\
24 18.78\\
25 18.8\\
26 18.7\\
27 18.84\\
28 18.88\\
29 17.74\\
30 18.74\\
31 18.84\\
32 19.98\\
33 18.86\\
34 18.48\\
35 18.84\\
36 18.84\\
37 18.86\\
38 18.86\\
39 18.84\\
40 18.76\\
41 19.98\\
42 18.8\\
43 17.54\\
44 19.98\\
45 19.98\\
46 18.8\\
47 19.98\\
48 18.64\\
49 18.7\\
50 17.24\\
51 18.82\\
52 18.84\\
53 18.78\\
54 18.68\\
55 19.98\\
56 18.8\\
57 18.82\\
58 16.84\\
59 17.36\\
60 16.98\\
61 18.72\\
62 18.8\\
63 18.58\\
64 18.66\\
65 18.88\\
66 19.98\\
67 19.98\\
68 19.98\\
69 18.86\\
70 17.16\\
71 18.56\\
72 19.98\\
73 18.86\\
74 16.36\\
75 18.9\\
76 17.56\\
77 18.44\\
78 19.98\\
79 18.88\\
80 17.7\\
81 18.84\\
82 19.98\\
83 18.84\\
84 18.84\\
85 18.66\\
86 15.76\\
87 18.1\\
88 18.88\\
89 19.98\\
90 17.64\\
91 18.9\\
92 19.98\\
93 18.8\\
94 19.98\\
95 19.98\\
96 18.9\\
97 18.8\\
98 19.98\\
99 19.98\\
};
\addlegendentry{$\text{SDP - }\beta{}_{\text{2}}\text{ =0}$};

\addplot [
color=blue,
dash pattern=on 1pt off 3pt on 3pt off 3pt,
line width=0.8pt
]
table[row sep=crcr]{
3 8.64\\
4 12.004\\
5 15.188\\
6 18.156\\
7 18.052\\
8 18.028\\
9 16.216\\
10 16.324\\
11 16.32\\
12 16.52\\
13 16.54\\
14 18.088\\
15 18.232\\
16 18.464\\
17 18.7\\
18 18.936\\
19 19.26\\
20 19.256\\
21 19.256\\
22 19.264\\
23 19.028\\
24 19.012\\
25 19.02\\
26 18.8\\
27 18.592\\
28 18.58\\
29 18.608\\
30 18.836\\
31 18.832\\
32 18.98\\
33 19\\
34 19\\
35 18.776\\
36 18.776\\
37 18.848\\
38 18.832\\
39 19.06\\
40 19.048\\
41 18.784\\
42 19.012\\
43 19.256\\
44 19.02\\
45 19.256\\
46 19.476\\
47 19.22\\
48 18.672\\
49 18.676\\
50 18.448\\
51 18.476\\
52 18.472\\
53 19.02\\
54 19.016\\
55 19.012\\
56 18.624\\
57 18.36\\
58 17.76\\
59 17.744\\
60 17.74\\
61 18.088\\
62 18.348\\
63 18.728\\
64 18.98\\
65 19.216\\
66 19.496\\
67 19.536\\
68 19.192\\
69 18.908\\
70 18.908\\
71 18.684\\
72 18.184\\
73 18.532\\
74 18.332\\
75 18.024\\
76 18.248\\
77 18.752\\
78 18.512\\
79 18.768\\
80 19.076\\
81 18.848\\
82 18.84\\
83 19.032\\
84 18.416\\
85 18.04\\
86 18.048\\
87 18.276\\
88 18.072\\
89 18.7\\
90 19.076\\
91 19.06\\
92 19.06\\
93 19.528\\
94 19.528\\
95 19.292\\
96 19.528\\
97 19.528\\
98 18.912\\
};
\addlegendentry{$\text{SDP - }\beta{}_{\text{2}}\text{ =0}$};
\addlegendimage{empty legend};
\addlegendentry{trend};

\addplot [
color=black,
line width=0.3pt,
mark size=1.5pt,
only marks,
mark=asterisk,
mark options={solid}
]
table[row sep=crcr]{
1 0.78\\
2 2.72\\
3 11.28\\
4 18.74\\
5 17.08\\
6 17.4\\
7 19.98\\
8 18.42\\
9 19.98\\
10 14.46\\
11 18.54\\
12 19.98\\
13 19.98\\
14 18.68\\
15 15.22\\
16 17.36\\
17 17.5\\
18 18.56\\
19 18.84\\
20 18.86\\
21 18.78\\
22 18.82\\
23 17.02\\
24 16.24\\
25 18.8\\
26 17.7\\
27 18.68\\
28 18.86\\
29 18.34\\
30 19.98\\
31 17.76\\
32 17.64\\
33 17.62\\
34 18.84\\
35 18.86\\
36 11.56\\
37 18.58\\
38 18.62\\
39 19.98\\
40 18.84\\
41 16.74\\
42 18.54\\
43 17.78\\
44 18.54\\
45 19.98\\
46 18.48\\
47 17.4\\
48 18.9\\
49 18.84\\
50 17.84\\
51 19.98\\
52 18.88\\
53 18.82\\
54 18.76\\
55 18.88\\
56 18.92\\
57 17.36\\
58 18.72\\
59 17.82\\
60 18.6\\
61 18.88\\
62 19.98\\
63 17.54\\
64 19.98\\
65 17.6\\
66 18.24\\
67 9.12\\
68 18.92\\
69 17.32\\
70 18.88\\
71 17.6\\
72 17.46\\
73 18.9\\
74 18.74\\
75 17.64\\
76 19.98\\
77 18.88\\
78 19.98\\
79 17.88\\
80 18.9\\
81 16.78\\
82 14.38\\
83 18.52\\
84 19.98\\
85 18.92\\
86 18.84\\
87 18.86\\
88 18.94\\
89 19.98\\
90 18.86\\
91 17.76\\
92 18.92\\
93 15.8\\
94 16.96\\
95 19.98\\
96 19.98\\
97 18.84\\
98 19.98\\
99 16.88\\
};
\addlegendentry{$\text{SDP - }\beta{}_{\text{2}}\text{ =0.4}$};

\addplot [
color=black,
solid,
line width=0.8pt
]
table[row sep=crcr]{
3 10.12\\
4 13.444\\
5 16.896\\
6 18.324\\
7 18.572\\
8 18.048\\
9 18.276\\
10 18.276\\
11 18.588\\
12 18.328\\
13 18.48\\
14 18.244\\
15 17.748\\
16 17.464\\
17 17.496\\
18 18.224\\
19 18.508\\
20 18.772\\
21 18.464\\
22 17.944\\
23 17.932\\
24 17.716\\
25 17.688\\
26 18.056\\
27 18.476\\
28 18.712\\
29 18.724\\
30 18.516\\
31 18.268\\
32 18.368\\
33 18.144\\
34 16.904\\
35 17.092\\
36 17.292\\
37 17.52\\
38 17.516\\
39 18.552\\
40 18.544\\
41 18.376\\
42 18.088\\
43 18.316\\
44 18.664\\
45 18.436\\
46 18.66\\
47 18.72\\
48 18.292\\
49 18.592\\
50 18.888\\
51 18.872\\
52 18.856\\
53 19.064\\
54 18.852\\
55 18.548\\
56 18.528\\
57 18.34\\
58 18.284\\
59 18.276\\
60 18.8\\
61 18.564\\
62 18.996\\
63 18.796\\
64 18.668\\
65 16.496\\
66 16.772\\
67 16.24\\
68 16.496\\
69 16.368\\
70 18.036\\
71 18.032\\
72 18.316\\
73 18.068\\
74 18.544\\
75 18.828\\
76 19.044\\
77 18.872\\
78 19.124\\
79 18.484\\
80 17.584\\
81 17.292\\
82 17.712\\
83 17.716\\
84 18.128\\
85 19.024\\
86 19.108\\
87 19.108\\
88 19.096\\
89 18.88\\
90 18.892\\
91 18.264\\
92 17.66\\
93 17.884\\
94 18.328\\
95 18.312\\
96 19.148\\
97 19.132\\
98 18.884\\
};
\addlegendentry{$\text{SDP - }\beta{}_{\text{2}}\text{ =0.4}$};
\addlegendimage{empty legend};
\addlegendentry{trend};

\end{axis}
\end{tikzpicture}%
}
	\caption{Performance overview of \acrshort{SDP} - $S_4^1 (\mathcal{T}_{32})$ ($480$ parameters), with ($\gls{forget}_{2} = 0.4$) and without ($\gls{forget}_{2} = 0$) forget factor, and with system noise ($\sigma_w = 3^{\circ}/s^2$). The trendline respresents a 5 point moving average.}
	\label{fig:e1_sdp_w3}
\end{figure*}

\subsection{Experiment II - The Time-Varying System} \label{sec:exp2}

The second experiment involves the control of a time-varying system. To simulate this, the control system has first been allowed to converge to the optimal value function by executing $1000$ learning trials. The pendulum's mass is then changed from $m = 1$ kg to $m = 1.5$ kg and then $100$ trials are simulated, similar to the first experiment.

\begin{figure*}
	\centering
%
%
%
\begin{tikzpicture}

\begin{axis}[%
width=\tikzwidth,
height=\tikzheight,
scale only axis,
xmin=0,
xmax=99,
xlabel={Trial},
xmajorgrids,
ymin=-0.1,
ymax=20.1,
ylabel={$t_{up}$ [s]},
ymajorgrids,
axis x line*=bottom,
axis y line*=left,
legend style={at={(1.03,0.5)},anchor=west,draw=black,fill=white,legend cell align=left}
]
\addplot [
color=blue,
line width=0.3pt,
mark size=1.5pt,
only marks,
mark=o,
mark options={solid}
]
table[row sep=crcr]{
1 2.62\\
2 3.7\\
3 0.5\\
4 0.52\\
5 0.94\\
6 0.46\\
7 1.04\\
8 0.42\\
9 2.18\\
10 0.58\\
11 0.5\\
12 2.34\\
13 3.84\\
14 4.3\\
15 1.22\\
16 0.88\\
17 0.74\\
18 0.56\\
19 0.58\\
20 0.8\\
21 0.64\\
22 0.74\\
23 0.88\\
24 0.78\\
25 0.84\\
26 0.84\\
27 0.82\\
28 0.82\\
29 0.92\\
30 9\\
31 0.96\\
32 0.96\\
33 15.68\\
34 18.9\\
35 18.8\\
36 19.98\\
37 18.48\\
38 18.64\\
39 19.98\\
40 18.82\\
41 16.26\\
42 18.44\\
43 17.3\\
44 18.54\\
45 17.72\\
46 16.08\\
47 14.76\\
48 16.38\\
49 18.78\\
50 17.16\\
51 19.98\\
52 18.82\\
53 16.5\\
54 16.06\\
55 16.5\\
56 16.46\\
57 17.44\\
58 18.64\\
59 17.72\\
60 18.56\\
61 16.6\\
62 19.98\\
63 16.74\\
64 19.98\\
65 17.46\\
66 17.9\\
67 18.56\\
68 18.88\\
69 17.42\\
70 18.66\\
71 16.66\\
72 17.62\\
73 16.02\\
74 18.68\\
75 17.52\\
76 19.98\\
77 15.98\\
78 19.98\\
79 16.92\\
80 16.12\\
81 17\\
82 17.56\\
83 18.4\\
84 19.98\\
85 18.82\\
86 18.78\\
87 16.04\\
88 18.88\\
89 19.98\\
90 18.76\\
91 17.56\\
92 18.72\\
93 18.76\\
94 16.58\\
95 19.98\\
96 19.98\\
97 18.58\\
98 19.98\\
99 17.14\\
};
\addlegendentry{NDP - FF};

\addplot [
color=blue,
dash pattern=on 1pt off 3pt on 3pt off 3pt,
line width=0.8pt
]
table[row sep=crcr]{
3 1.656\\
4 1.224\\
5 0.692\\
6 0.676\\
7 1.008\\
8 0.936\\
9 0.944\\
10 1.204\\
11 1.888\\
12 2.312\\
13 2.44\\
14 2.516\\
15 2.196\\
16 1.54\\
17 0.796\\
18 0.712\\
19 0.664\\
20 0.664\\
21 0.728\\
22 0.768\\
23 0.776\\
24 0.816\\
25 0.832\\
26 0.82\\
27 0.848\\
28 2.48\\
29 2.504\\
30 2.532\\
31 5.504\\
32 9.1\\
33 11.06\\
34 14.864\\
35 18.368\\
36 18.96\\
37 19.176\\
38 19.18\\
39 18.436\\
40 18.428\\
41 18.16\\
42 17.872\\
43 17.652\\
44 17.616\\
45 16.88\\
46 16.696\\
47 16.744\\
48 16.632\\
49 17.412\\
50 18.224\\
51 18.248\\
52 17.704\\
53 17.572\\
54 16.868\\
55 16.592\\
56 17.02\\
57 17.352\\
58 17.764\\
59 17.792\\
60 18.3\\
61 17.92\\
62 18.372\\
63 18.152\\
64 18.412\\
65 18.128\\
66 18.556\\
67 18.044\\
68 18.284\\
69 18.036\\
70 17.848\\
71 17.276\\
72 17.528\\
73 17.3\\
74 17.964\\
75 17.636\\
76 18.428\\
77 18.076\\
78 17.796\\
79 17.2\\
80 17.516\\
81 17.2\\
82 17.812\\
83 18.352\\
84 18.708\\
85 18.404\\
86 18.5\\
87 18.5\\
88 18.488\\
89 18.244\\
90 18.78\\
91 18.756\\
92 18.076\\
93 18.32\\
94 18.804\\
95 18.776\\
96 19.02\\
97 19.132\\
98 18.624\\
};
\addlegendentry{NDP - FF};
\addlegendimage{empty legend};
\addlegendentry{trend};

\addplot [
color=black,
line width=0.3pt,
mark size=1.5pt,
only marks,
mark=asterisk,
mark options={solid}
]
table[row sep=crcr]{
1 1.2\\
2 0\\
3 0.72\\
4 0\\
5 1\\
6 1.2\\
7 0.78\\
8 0.56\\
9 1.62\\
10 1.9\\
11 1.94\\
12 1.3\\
13 3.46\\
14 2.08\\
15 1.7\\
16 1.58\\
17 0.92\\
18 0.8\\
19 1\\
20 1.1\\
21 1.16\\
22 1.22\\
23 0.94\\
24 1.48\\
25 1.18\\
26 1.4\\
27 0.94\\
28 1.3\\
29 1.22\\
30 1.76\\
31 0\\
32 1.44\\
33 1.4\\
34 1.4\\
35 1.46\\
36 1.7\\
37 1.74\\
38 1.64\\
39 1.62\\
40 1.7\\
41 1.38\\
42 1.52\\
43 1.44\\
44 1.36\\
45 1.76\\
46 1.56\\
47 1.72\\
48 1.68\\
49 1.62\\
50 1.64\\
51 1.62\\
52 1.68\\
53 1.7\\
54 1.52\\
55 1.78\\
56 1.8\\
57 1.58\\
58 1.48\\
59 1.62\\
60 1.36\\
61 1.26\\
62 1.64\\
63 1.66\\
64 2.2\\
65 2.2\\
66 1.7\\
67 1.64\\
68 1.9\\
69 1.86\\
70 1.88\\
71 1.84\\
72 1.82\\
73 1.84\\
74 1.7\\
75 1.7\\
76 2.32\\
77 2.5\\
78 2.86\\
79 3.02\\
80 3.3\\
81 2.78\\
82 2.64\\
83 2.14\\
84 3.88\\
85 3.56\\
86 3.54\\
87 4\\
88 4.36\\
89 5.9\\
90 5.34\\
91 6.28\\
92 7.46\\
93 7.94\\
94 10.54\\
95 19.98\\
96 19.98\\
97 18.88\\
98 19.98\\
99 17.74\\
};
\addlegendentry{NDP - RBF};

\addplot [
color=black,
solid,
line width=0.8pt
]
table[row sep=crcr]{
3 0.584\\
4 0.584\\
5 0.74\\
6 0.708\\
7 1.032\\
8 1.212\\
9 1.36\\
10 1.464\\
11 2.044\\
12 2.136\\
13 2.096\\
14 2.024\\
15 1.948\\
16 1.416\\
17 1.2\\
18 1.08\\
19 0.996\\
20 1.056\\
21 1.084\\
22 1.18\\
23 1.196\\
24 1.244\\
25 1.188\\
26 1.26\\
27 1.208\\
28 1.324\\
29 1.044\\
30 1.144\\
31 1.164\\
32 1.2\\
33 1.14\\
34 1.48\\
35 1.54\\
36 1.588\\
37 1.632\\
38 1.68\\
39 1.616\\
40 1.572\\
41 1.532\\
42 1.48\\
43 1.492\\
44 1.528\\
45 1.568\\
46 1.616\\
47 1.668\\
48 1.644\\
49 1.656\\
50 1.648\\
51 1.652\\
52 1.632\\
53 1.66\\
54 1.696\\
55 1.676\\
56 1.632\\
57 1.652\\
58 1.568\\
59 1.46\\
60 1.472\\
61 1.508\\
62 1.624\\
63 1.792\\
64 1.88\\
65 1.88\\
66 1.928\\
67 1.86\\
68 1.796\\
69 1.824\\
70 1.86\\
71 1.848\\
72 1.816\\
73 1.78\\
74 1.876\\
75 2.012\\
76 2.216\\
77 2.48\\
78 2.8\\
79 2.892\\
80 2.92\\
81 2.776\\
82 2.948\\
83 3\\
84 3.152\\
85 3.424\\
86 3.868\\
87 4.272\\
88 4.628\\
89 5.176\\
90 5.868\\
91 6.584\\
92 7.512\\
93 10.44\\
94 13.18\\
95 15.464\\
96 17.872\\
97 19.312\\
98 18.856\\
};
\addlegendentry{NDP - RBF};
\addlegendimage{empty legend};
\addlegendentry{trend};

\end{axis}
\end{tikzpicture}%
	\caption{Performance overview of \acrshort{NDP} - Radial basis 12x12 ($432$ parameters) and feedforward network 160 neurons ($480$ parameters), with an altered system parameter ($m = 1.5$) at $t = 0$ and without system noise ($\sigma_w = 0^{\circ}/s^2$). The trendline respresents a 5 point moving average.}
	\label{fig:e2_trial_ndp}
\end{figure*}
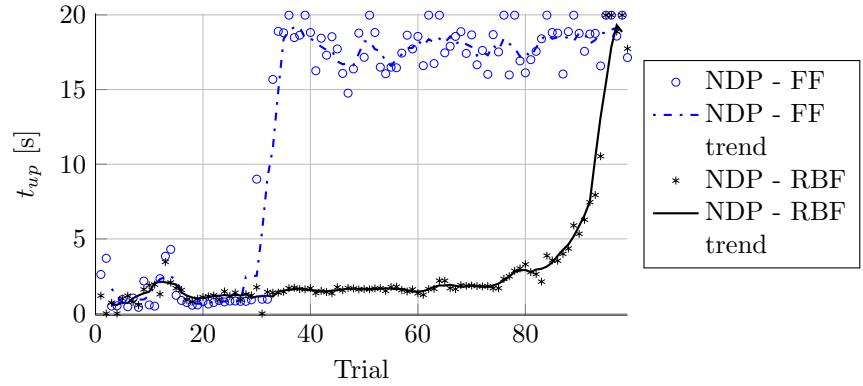

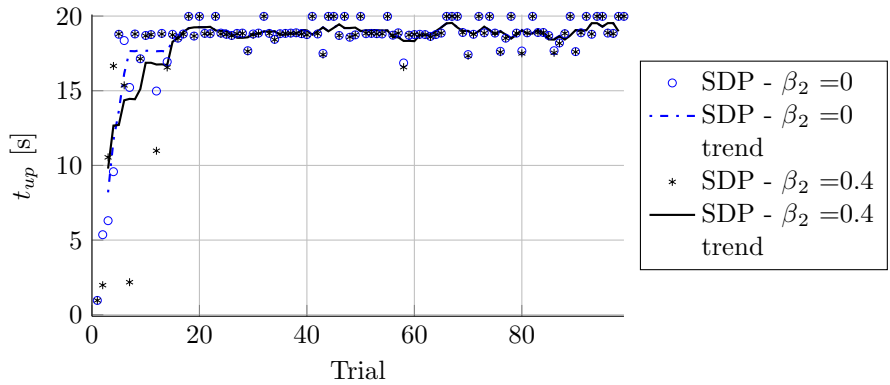
\begin{figure*}
	\centering
%
%
%
\begin{tikzpicture}

\begin{axis}[%
width=\tikzwidth,
height=\tikzheight,
scale only axis,
xmin=0,
xmax=99,
xlabel={Trial},
xmajorgrids,
ymin=-0.1,
ymax=20.1,
ylabel={$t_{up}$ [s]},
ymajorgrids,
axis x line*=bottom,
axis y line*=left,
legend style={at={(1.03,0.5)},anchor=west,draw=black,fill=white,legend cell align=left}
]
\addplot [
color=blue,
line width=0.3pt,
mark size=1.5pt,
only marks,
mark=o,
mark options={solid}
]
table[row sep=crcr]{
1 0.96\\
2 5.36\\
3 6.3\\
4 9.58\\
5 18.78\\
6 18.36\\
7 15.22\\
8 18.8\\
9 17.14\\
10 18.7\\
11 18.76\\
12 14.98\\
13 18.84\\
14 16.94\\
15 18.78\\
16 18.52\\
17 18.8\\
18 19.98\\
19 18.66\\
20 19.98\\
21 18.84\\
22 18.82\\
23 19.98\\
24 18.86\\
25 18.78\\
26 18.7\\
27 18.84\\
28 18.86\\
29 17.68\\
30 18.8\\
31 18.86\\
32 19.98\\
33 18.84\\
34 18.44\\
35 18.82\\
36 18.84\\
37 18.86\\
38 18.88\\
39 18.82\\
40 18.76\\
41 19.98\\
42 18.8\\
43 17.5\\
44 19.98\\
45 19.98\\
46 18.7\\
47 19.98\\
48 18.58\\
49 18.74\\
50 19.98\\
51 18.82\\
52 18.84\\
53 18.82\\
54 18.8\\
55 19.98\\
56 18.72\\
57 18.88\\
58 16.86\\
59 18.72\\
60 18.76\\
61 18.78\\
62 18.76\\
63 18.64\\
64 18.76\\
65 18.86\\
66 19.98\\
67 19.98\\
68 19.98\\
69 18.9\\
70 17.42\\
71 18.78\\
72 19.98\\
73 18.88\\
74 19.98\\
75 18.86\\
76 17.64\\
77 18.52\\
78 19.98\\
79 18.86\\
80 17.66\\
81 18.88\\
82 19.98\\
83 18.9\\
84 18.9\\
85 18.7\\
86 17.68\\
87 18.2\\
88 18.82\\
89 19.98\\
90 17.62\\
91 18.88\\
92 19.98\\
93 18.78\\
94 19.98\\
95 19.98\\
96 18.86\\
97 18.84\\
98 19.98\\
99 19.98\\
};
\addlegendentry{$\text{SDP - }\beta{}_{\text{2}}\text{ =0}$};

\addplot [
color=blue,
dash pattern=on 1pt off 3pt on 3pt off 3pt,
line width=0.8pt
]
table[row sep=crcr]{
3 8.196\\
4 11.676\\
5 13.648\\
6 16.148\\
7 17.66\\
8 17.644\\
9 17.724\\
10 17.676\\
11 17.684\\
12 17.644\\
13 17.66\\
14 17.612\\
15 18.376\\
16 18.604\\
17 18.948\\
18 19.188\\
19 19.252\\
20 19.256\\
21 19.256\\
22 19.296\\
23 19.056\\
24 19.028\\
25 19.032\\
26 18.808\\
27 18.572\\
28 18.576\\
29 18.608\\
30 18.836\\
31 18.832\\
32 18.984\\
33 18.988\\
34 18.984\\
35 18.76\\
36 18.768\\
37 18.844\\
38 18.832\\
39 19.06\\
40 19.048\\
41 18.772\\
42 19.004\\
43 19.248\\
44 18.992\\
45 19.228\\
46 19.444\\
47 19.196\\
48 19.196\\
49 19.22\\
50 18.992\\
51 19.04\\
52 19.052\\
53 19.052\\
54 19.032\\
55 19.04\\
56 18.648\\
57 18.632\\
58 18.388\\
59 18.4\\
60 18.376\\
61 18.732\\
62 18.74\\
63 18.76\\
64 19\\
65 19.244\\
66 19.512\\
67 19.54\\
68 19.252\\
69 19.012\\
70 19.012\\
71 18.792\\
72 19.008\\
73 19.296\\
74 19.068\\
75 18.776\\
76 18.996\\
77 18.772\\
78 18.532\\
79 18.78\\
80 19.072\\
81 18.856\\
82 18.864\\
83 19.072\\
84 18.832\\
85 18.476\\
86 18.46\\
87 18.676\\
88 18.46\\
89 18.7\\
90 19.056\\
91 19.048\\
92 19.048\\
93 19.52\\
94 19.516\\
95 19.288\\
96 19.528\\
97 19.528\\
98 18.996\\
};
\addlegendentry{$\text{SDP - }\beta{}_{\text{2}}\text{ =0}$};
\addlegendimage{empty legend};
\addlegendentry{trend};

\addplot [
color=black,
line width=0.3pt,
mark size=1.5pt,
only marks,
mark=asterisk,
mark options={solid}
]
table[row sep=crcr]{
1 0.96\\
2 1.98\\
3 10.54\\
4 16.66\\
5 18.8\\
6 15.36\\
7 2.18\\
8 18.8\\
9 17.14\\
10 18.68\\
11 18.76\\
12 10.98\\
13 18.8\\
14 16.56\\
15 18.74\\
16 18.52\\
17 18.78\\
18 19.98\\
19 18.66\\
20 19.98\\
21 18.82\\
22 18.82\\
23 19.98\\
24 18.8\\
25 18.76\\
26 18.68\\
27 18.82\\
28 18.84\\
29 17.64\\
30 18.74\\
31 18.84\\
32 19.98\\
33 18.84\\
34 18.42\\
35 18.8\\
36 18.84\\
37 18.84\\
38 18.86\\
39 18.8\\
40 18.74\\
41 19.98\\
42 18.82\\
43 17.42\\
44 19.98\\
45 19.98\\
46 18.7\\
47 19.98\\
48 18.6\\
49 18.74\\
50 19.98\\
51 18.84\\
52 18.84\\
53 18.8\\
54 18.82\\
55 19.98\\
56 18.72\\
57 18.86\\
58 16.58\\
59 18.74\\
60 18.78\\
61 18.74\\
62 18.76\\
63 18.6\\
64 18.76\\
65 18.86\\
66 19.98\\
67 19.98\\
68 19.98\\
69 18.9\\
70 17.34\\
71 18.78\\
72 19.98\\
73 18.86\\
74 19.98\\
75 18.86\\
76 17.58\\
77 18.52\\
78 19.98\\
79 18.86\\
80 17.5\\
81 18.88\\
82 19.98\\
83 18.9\\
84 18.9\\
85 18.66\\
86 17.54\\
87 18.18\\
88 18.84\\
89 19.98\\
90 17.6\\
91 18.88\\
92 19.98\\
93 18.8\\
94 19.98\\
95 19.98\\
96 18.88\\
97 18.82\\
98 19.98\\
99 19.98\\
};
\addlegendentry{$\text{SDP - }\beta{}_{\text{2}}\text{ =0.4}$};

\addplot [
color=black,
solid,
line width=0.8pt
]
table[row sep=crcr]{
3 9.788\\
4 12.668\\
5 12.708\\
6 14.36\\
7 14.456\\
8 14.432\\
9 15.112\\
10 16.872\\
11 16.872\\
12 16.756\\
13 16.768\\
14 16.72\\
15 18.28\\
16 18.516\\
17 18.936\\
18 19.184\\
19 19.244\\
20 19.252\\
21 19.252\\
22 19.28\\
23 19.036\\
24 19.008\\
25 19.008\\
26 18.78\\
27 18.548\\
28 18.544\\
29 18.576\\
30 18.808\\
31 18.808\\
32 18.964\\
33 18.976\\
34 18.976\\
35 18.748\\
36 18.752\\
37 18.828\\
38 18.816\\
39 19.044\\
40 19.04\\
41 18.752\\
42 18.988\\
43 19.236\\
44 18.98\\
45 19.212\\
46 19.448\\
47 19.2\\
48 19.2\\
49 19.228\\
50 19\\
51 19.04\\
52 19.056\\
53 19.056\\
54 19.032\\
55 19.036\\
56 18.592\\
57 18.576\\
58 18.336\\
59 18.34\\
60 18.32\\
61 18.724\\
62 18.728\\
63 18.744\\
64 18.992\\
65 19.236\\
66 19.512\\
67 19.54\\
68 19.236\\
69 18.996\\
70 18.996\\
71 18.772\\
72 18.988\\
73 19.292\\
74 19.052\\
75 18.76\\
76 18.984\\
77 18.76\\
78 18.488\\
79 18.748\\
80 19.04\\
81 18.824\\
82 18.832\\
83 19.064\\
84 18.796\\
85 18.436\\
86 18.424\\
87 18.64\\
88 18.428\\
89 18.696\\
90 19.056\\
91 19.048\\
92 19.048\\
93 19.524\\
94 19.524\\
95 19.292\\
96 19.528\\
97 19.528\\
98 18.98\\
};
\addlegendentry{$\text{SDP - }\beta{}_{\text{2}}\text{ =0.4}$};
\addlegendimage{empty legend};
\addlegendentry{trend};

\end{axis}
\end{tikzpicture}%
	\caption{Performance overview of \acrshort{SDP} - $S_4^1 (\mathcal{T}_{32})$ ($480$ parameters), with ($\gls{forget}_{2} = 0.4$) and without ($\gls{forget}_{2} = 0$) forget factor, with an altered system parameter ($m = 1.5$) at $t = 0$ and without system noise ($\sigma_w = 0^{\circ}/s^2$). The trendline respresents a 5 point moving average.}
	\label{fig:e2_trial_sdp}
\end{figure*}

The $t_{up}$ of $100$ trials after the change are visible in figures~\ref{fig:e2_trial_ndp} and \ref{fig:e2_trial_sdp} for the \gls{NDP} and \gls{SDP} methods respectively. 
By increasing the pendulum's mass by $50\%$ the old control system does not stop working, in fact, in many cases the top can still be reached, albeit not as close to $\gls{theta} = 0$ as before the change (i.e. the pendulum is held stationary at a slight angle). 
The changed optimal value function introduces a \acrshort{TD}-error which propagates through the network.
For the feedforward network a minor adaptation of the parameters is sufficient to adapt the global shape of the estimated value function, allowing the \acrshort{NDP} - feedforward to continue controlling the altered system without temporary decreased performance. This is in contrast to \acrshort{NDP} - radial basis, which does produce a decrease in performance as the \acrshort{TD}-error propagates through the estimated value function.
\acrshort{SDP} - $\gls{forget}_{2} = 0$ gives each data-point an equal weight, it is slow in adapting itself to a new situation, spending a long period in the transition phase where performance is reduced. \acrshort{SDP} - $\gls{forget}_{2} = 0.4$ has a shorter transition phase and has adapted itself to the new system before the $50^{th}$ trial.

In Table~\ref{tab:results2}, the mean and standard deviation of $t_{up}$ of the $100$ trials are presented. This quantification identifies \acrshort{NDP} - feedforward as the method with the best performance, and \acrshort{SDP} - $\gls{forget}_{2} = 0.4$ as the method with the second best performance. As stated before, the performance of the feedforward network is a direct consequence of its ability to generalize. 
However, there is no guarantee of convergence. 
Furthermore, it identifies \acrshort{NDP} - radial basis and \acrshort{SDP} - $\gls{forget}_{2} = 0$ as equally bad, however in the figure~\ref{fig:e2_trial_ndp} it can be seen that \acrshort{NDP} - radial basis has recovered from the system change in the last $20$ trials. This indicates that \acrshort{NDP} - radial basis is capable of recovering although it takes more trials than the other methods.

\begin{table}
\caption{Mean and standard deviation (std) of $t_{up}$ of the $100$ trials of Experiment II}
\label{tab:results2}
\centering
    \begin{tabular}{c|c|c|c|c}    
	\hline
	\hline
			& \acrshort{NDP} - 	& \acrshort{NDP} - 	& \acrshort{SDP} -  	& \acrshort{SDP} -  \\
			& FF 		& RBF 		& $\gls{forget}_{2} = 0$ & $\gls{forget}_{2} = 0.4$ \\
	  \hline
	  \textbf{Mean}	& $17.88$ s	& $11.44$ s	& $12.40$ s	& $15.90$ s	\\
	  \textbf{Std}	& $1.47$ s	& $7.89$ s	& $6.76$ s	& $5.15$ s	\\
	\hline
	\hline
    \end{tabular}
\end{table}


\section{Discussion} \label{sec:discussion} 
The most important difference between neural networks and simplex splines for a \gls{DP} framework, is that neural networks are non-linear in the parameters, while simplex splines are linear-in-the-parameters. Using a linear-in-the-parameters function approximator allows for the use of the \gls{RLSTD} algorithm, which has a fast and proven convergence in a stochastic framework \citea{BB96}.
As a result, the \gls{SDP} framework without forget factor ($\gls{forget}_{2} = 0$) has proven convergence demonstrates stable performance when learning online. Nevertheless, in certain circumstances it is beneficial to trade these properties for adaptability, which is done by using the modified \gls{RLSTD} algorithm from Table~\ref{tab:trainRLSTDmod} and thus introducing a forget factor ($\gls{forget}_{2} > 0$). These circumstances arise when the environment is susceptible to system parameter changes and quick adaptation is required.

To successfully implement the \gls{SDP} framework, it is important to construct the proper spline space.
Because, even though \gls{SDP} has proven convergence to the best fit, there is no guarantee of system performance. To obtain this guarantee, the optimal value function, or its shape, must be known a-priori. In practice this will mean that either an off-line simulation is used, or the system is tested to see if the desired performance is reached.
Another option is to treat all unknown parameters as an additional optimization, and solve the entire optimization problem using Intersplines \citea{dVvK+12}. Unfortunately, at the moment Intersplines are limited to two-dimensional inputs, and require too much calculation power to make it attractive for real-time applications.

While it is possible to use the multivariate simplex B-splines in higher dimensions \citea{Boor1986}, there are two problems that arise. First there is the construction of the triangulation, which is not automated and is already tedious in a dimension $N = 3$. Secondly, due to the ``curse of dimensionality'' \citea{Bellman1957}, the computational costs of dynamic programming are very high when moving to higher dimensions. The effects of this curse are reduced by selecting a triangulation that has fewer simplices, but retains the ability to represent the optimal value function. However, the first argument was that constructing a triangulation becomes increasingly difficult at higher dimensions and is not yet automated. This creates difficulties when applying the \gls{SDP} framework to a problem with the dimension $N>2$, since a similar grid-approach as used in the simulation, produces high computation costs.

A final remark on the methods used in the experiments. The four methods were considered to be comparable in terms of the number of parameters. However, the continuity constraints reduced the amount of free parameters in the \gls{SDP} methods; thereby reducing the approximation power of the splines relative to the neural networks. Experiments with \gls{NDP} have shown that reducing the amount of parameters reduces the performance, which implies that \gls{SDP} is potentially better than concluded in the experiments.

\section{Conclusion} \label{sec:conclusion} 
In this paper the \gls{SDP} framework was introduced; a combination of the \gls{RLSTD} algorithm and multivariate simplex B-splines. It was shown that it is capable of solving the non-linear control problem of the pendulum swing-up with nothing but a reward function as feedback. This was done in significantly less trials than \gls{NDP} systems supported by function approximators with a comparable amount of parameters.
In addition, \gls{SDP} presented a greater resilience to system noise than \gls{NDP}, demonstrating no decrease in performance in the presence of a disturbance.
Furthermore, a forget method that preserves the continuity constraints is introduced, which is merged with the \gls{RLSTD} algorithm to create an adaptive control system.
In conclusion it can be said that the high convergence rate of the \gls{RLSTD} algorithm, in combination with the high approximation power of the multivariate simplex B-splines, provides a basis for high performance non-linear control at the cost of a higher computational load requirement. By using the modified \gls{RLSTD} algorithm, it is even possible to keep track of time-varying systems, resulting in an adaptive control method for non-linear systems.


In order to have a good trade-off between noise filtering and adaptability, it is important to design a forget factor that is capable of detecting a parameter change in the controlled system. There is extensive literature available on this subject and it is recommended to investigate which is the most effective in the \gls{SDP} framework.

The unsolved issue of multivariate simplex B-splines remains the search for the optimal triangulation. While the assumed static triangulation performs adequately, as seen in the experiments, much can be gained in terms of computational complexity and performance by optimizing the triangulation. The advantage of finding the optimal triangluation grows exponentially, as it is connected to the ``Curse of Dimensionality''.


\appendix

\section{Pendulum Swing-Up Task} \label{sec:swingup}

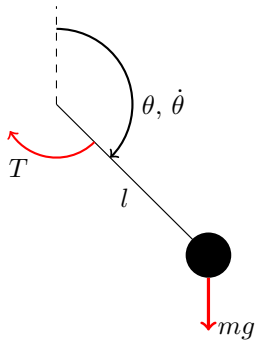
\begin{figure}
\centering
  \begin{tikzpicture}
  \draw [dashed] (0,0) -- (0,1.3);
  \draw [->,thick] (0,1) arc (90:-45:1); \draw node at (1,0) [right] {$\theta$, $\dot \theta$};
  \fill [black] (0,0) -- (2,-2) circle [radius=.3];
  \draw (0,0) -- (2,-2);
  \draw [->,very thick,red] (2,-2.3) -- (2,-3); \draw node at (2,-3) [right] {$m g$};
  \draw [->,thick,red] (.5,-.5) arc (-45:-150:.707); \draw node at (-.5,-.6) [below] {$T$};
  \draw node at (.9,-1) [below] {$l$};
  \end{tikzpicture}
\caption{Pendulum Swing-Up}
\label{fig:pen}
\end{figure}

The dynamics of the non-linear control task determined by:
\begin{equation} ml^2\gls{thetadd} = - \gls{mu} \gls{thetad} + mgl \sin{\gls{theta}} + T \end{equation}
where \gls{theta} is the angle from upright position, $T$ is the limited input torque, $\gls{mu} = 0.01$ is the friction coefficient, $m  = 1.0 kg$ is the point mass at the tip of the rod, $l = 1.0 m$ is the length of the pendulum, and $g = 9.8 m/s^2$ is the gravity acceleration. A schematic overview is visible in Fig.~\ref{fig:pen}.

The state vector is defined as $\gls{x} = [\gls{theta} ~ \gls{thetad}]^{\top}$ and the action as $\gls{u} = T$. 
The complete system description is now:
\begin{equation} \gls{xd} = f(\gls{x},\gls{u}) = \left( \begin{array}{c} \gls{thetad} \\ \frac{g}{l} \sin{\gls{theta}} - \frac{\mu}{ml^2}\gls{thetad} \end{array} \right) + \left( \begin{array}{c} 0 \\ \frac{1}{ml^2} \end{array} \right) \gls{u} + \left( \begin{array}{c} 0 \\ 1 \end{array} \right) w \end{equation}
where $w$ represents white noise with a standard deviation of $\sigma_w$. 
The reward function of the system is:
\begin{equation} r(\gls{x},\gls{u}) = c_x (\cos{\gls{theta}} -1) + c_u \int_0^{\frac{\gls{T}}{\gls{T}^{\max}}} \tan( \frac{\pi}{2} s) ds  \end{equation}
where $c_x = 1$, $c_u = 0.1$ and $T^{\max} = 5$.
As determined in section~\ref{sec:pi}, the optimal control law is:
\begin{equation} \gls{u} = T^{\max} \tanh( \frac{\pi}{2} \frac{1}{\gls{cj}} \gls{tau} \frac{\partial \gls{V}^*}{\partial \gls{x}} \frac{\partial f(\gls{x},\gls{u})}{\partial \gls{u}} + n ) \label{eq:policy} \end{equation}
where $n$ represents white noise with a standard deviation of $\sigma_n = 0.01$, used as an explorer.


\bibliographystyle{abbrvnat}
\bibliography{references}

\end{document}